\documentclass{article}

 \usepackage[logo]{ace_ego}

\usepackage[utf8]{inputenc}
\usepackage{url}
\usepackage{amsfonts}
\usepackage{nicefrac}
\usepackage{algorithm}
\usepackage{algorithmic}
\usepackage{epsfig}
\usepackage{amsmath}
\usepackage{amssymb}
\usepackage{array}
\usepackage{tikz}
\usepackage{comment}
\usepackage{pifont}
\usepackage{listings}
\usepackage{marvosym}
\title{\raisebox{-0.2em}{\includegraphics[height=1.2em]{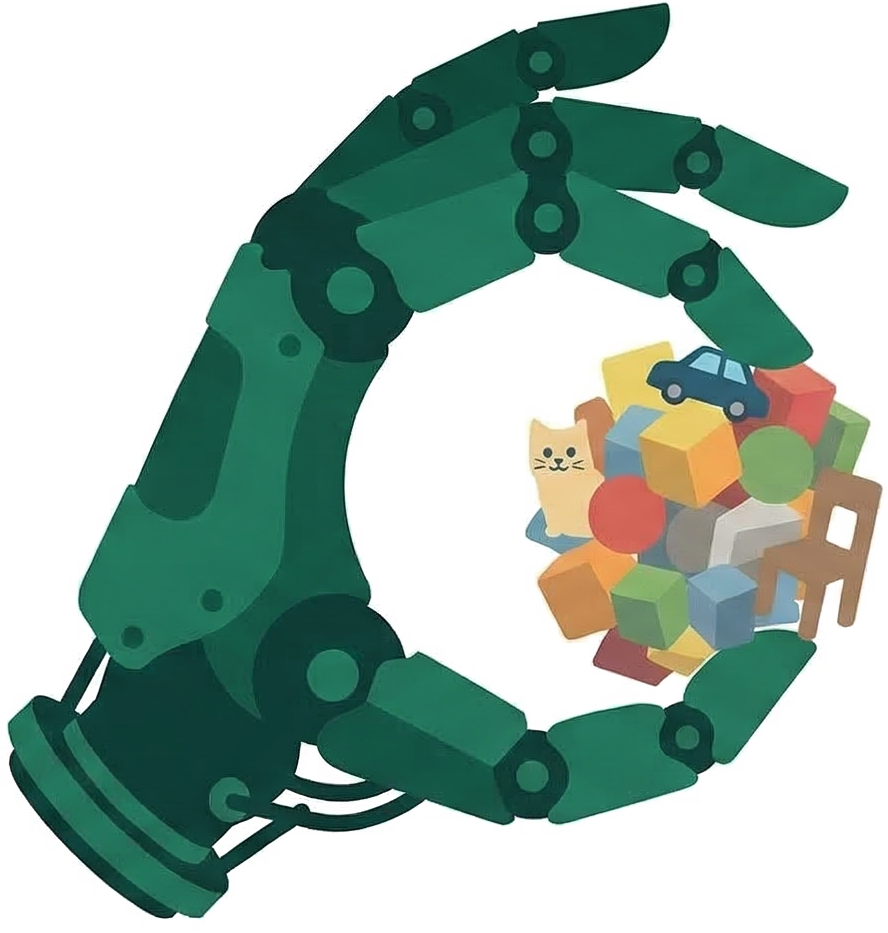}}mniDex: Scaling Dexterous Hand Grasping to Diverse Cluttered Scenes}

\author{%
  \sffamily\bfseries\small
  Naiyu Fang$^{1,2,3*}$ \And
  Zhongjin Luo$^{1,2,3*\,\dagger}$ \And
  Yuxin Mo$^{1,4}$ \And
  Siyuan Huang$^{1\,\dagger}$ \And
  Jianbo Liu$^{1}$ \And
  Yufei Liu$^{1,5}$ \AND
  \sffamily\bfseries\small
  Zheyuan Zhou$^{6}$ \And
  Chengkai Jin$^{1,7}$ \And
  Xiaogang Wang$^{1,2}$ \And
  Hongsheng Li$^{2{\,\scalebox{0.75}{\normalfont\Letter}}}$
  \\[6pt]
  \mdseries\normalsize
  \begin{tabular}[t]{@{}l@{}}
    $^{1}$ACE Robotics \quad $^{2}$CUHK, MMLab \quad $^{3}$CPII under InnoHK \quad $^{4}$Tongji University\\
    $^{5}$Shanghai Jiao Tong University \quad $^{6}$Zhejiang University \quad $^{7}$Nanyang Technological University\\[3pt]
    {\footnotesize $^*$Equal contribution \quad $^\dagger$Project lead \quad $^{{\scalebox{0.75}{\normalfont\Letter}}}$Corresponding author}
  \end{tabular}
}

\metadata[Website]{\url{https://aceroboticsdex.github.io/OmniDex/}}
\date{September 2026}

\begin{document}

\maketitle

\begin{abstract}
Dexterous grasping is the foundational primitive in embodied AI, demanding massive data to train robust models. As real-world data collection is expensive, simulation has become the mainstream paradigm. Yet, while cluttered scenes best reflect real-world applications, learning to grasp within them is bottlenecked by a critical scarcity of large-scale data. To resolve this, we curate high-quality 3D objects and supporting bases, proposing a scalable seed-and-filter strategy that bypasses sluggish scene-level optimization. This yields an unprecedented benchmark comprising over 2.6 million scenes and 0.4B scene-specific grasp ground truths, featuring diverse realistic layouts paired with rich semantic and geometric observations. Furthermore, we introduce the OmniDex model to overcome the grasp multimodality and last-millimeter precision errors plaguing current generative models. By coupling Soft Winner-Takes-All learning with human-inspired physical constraints during training, and utilizing physics-driven ranking, our approach achieves robust dexterous grasping without the latency of post-optimization. Experimental results show that OmniDex model achieves state-of-the-art performance and strong generalization across diverse scenes, views, and unseen objects.
\end{abstract}

\begin{figure}[htbp]
  \centering
  \includegraphics[width=\linewidth]{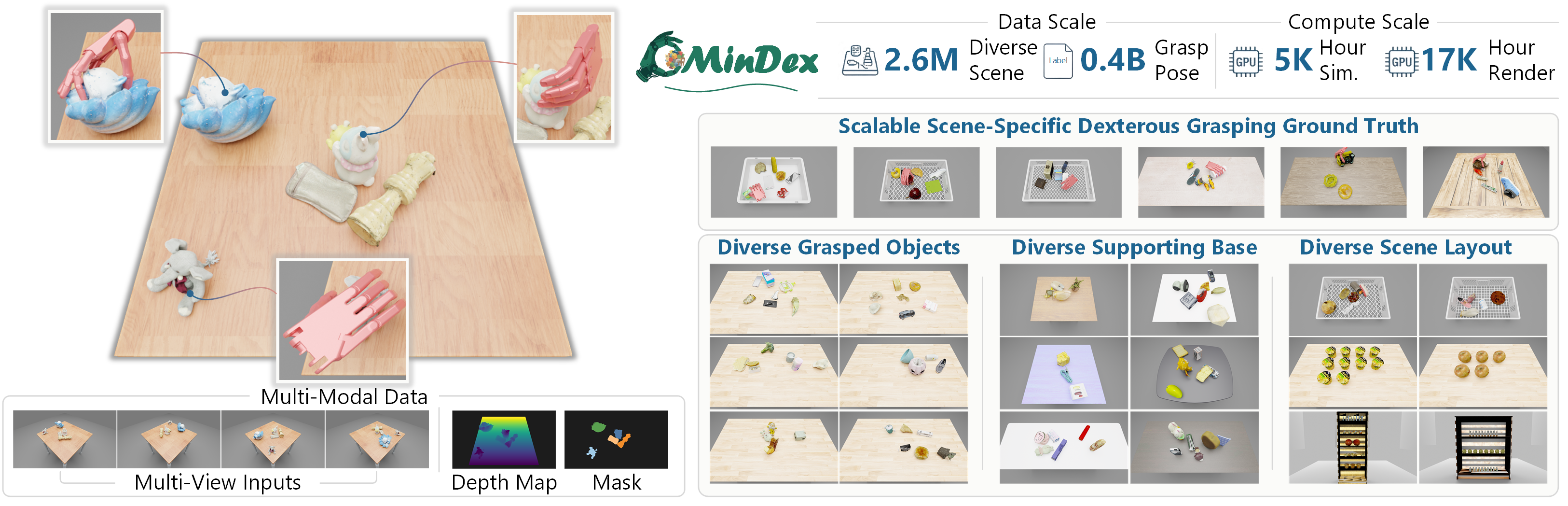}
  \caption{\textbf{Overview of OmniDex}, the largest benchmark for dexterous grasping in cluttered scenes to date. \textbf{Left}: Multi-view RGB-D and instruction masks paired with scene-specific grasp poses. \textbf{Right}: A scalable seed-and-filter strategy enables unprecedented scale (2.6M scenes, 0.4B grasp ground truths), encompassing diverse objects, supporting bases, and realistic layouts.}
  \label{fig0}
\end{figure}

\section{Introduction}
\label{sec1}

Dexterous hands mark a critical milestone in embodied artificial intelligence (AI) \cite{duan2022survey}. Unlike parallel-jaw grippers \cite{fang2020graspnet,fang2023anygrasp}, they offer the kinematic flexibility and fine-grained manipulation essential for complex, human-centric tasks. At the core of this capability lies grasping—the foundational primitive upon which all advanced interactions are built. For an intelligent agent, predicting a valid grasp pose acts as the "key frame" that dictates the success of all subsequent tasks. However, training grasp pose generation models is severely bottlenecked by a profound scarcity of high-quality data. Since collecting high degree-of-freedom (DoF) grasp poses in the real world is expensive and unscalable \cite{liu2024realdex}, simulation-based data synthesis \cite{yedex1b, chendexonomy} has emerged as the mainstream paradigm. To drive dexterous grasping task toward greater robustness and real-world applicability, this paper introduces a large-scale simulation benchmark targeting cluttered scenes, coupled with a novel learning framework.

Real-world dexterous grasping is inherently messy and cluttered. Yet, the standard setting in prior dexterous grasping literature oversimplifies the problem by isolating a single object in floating space or on a clear tabletop \cite{wang2023dexgraspnet, xu2023unidexgrasp}. Recent efforts tackle cluttered scenes \cite{zhang2024dexgraspnet, zhangcadgrasp}; however, robust performance in complex real-world settings remains an open problem due to three underlying challenges. First, current approaches rely on depth-based pseudo point clouds as their primary perceptual input. This forces models to focus on geometric shapes at the expense of semantic awareness. Consequently, agents struggle with target object localization in cluttered scenes (e.g., distinguishing geometrically identical Coke and Sprite cans), since their geometries are nearly identical. Second, existing benchmark scales are critically constrained by computationally expensive data generation pipelines. Scene-level optimization for grasp ground truths \cite{zhang2024dexgraspnet} incurs prohibitive time costs, limiting existing cluttered-scene benchmarks to only 0.4K--8K scenes (e.g., DexGraspNet 2.0: 8K; Tab.~\ref{table1}). Yet, massive scale is indispensable. Unlike single-object floating setups, cluttered scenes feature complex spatial dynamics among object. Thus, models require a large-scale dataset to effectively learn these spatial relationships and master scene-specific grasping, such as avoiding adjacent items or table penetration. Finally, prior works neglect diverse supporting bases and spatial layouts, compromising robustness and real-world generalization.

Beyond the data bottleneck, existing dexterous grasping generation models face critical algorithmic hurdles when mapping perceptual inputs to high-DoF kinematic space. Dexterous grasping is a multimodal problem, as an object affords numerous viable grasp poses. To capture this complex distribution, data-driven approaches rely heavily on intermediate representations, such as affordance maps \cite{zhang2024dexgraspnet, he2025dexvlg}, rendering the generation pipeline inherently non-end-to-end. However, pure data-driven methods struggle with critical last-millimeter precision errors. A mere millimeter of deviation translates directly to catastrophic contact failures in simulation. To compensate for this inherent inaccuracy, current methods resort to costly post-optimization or test-time fine-tuning \cite{wei2025afforddexgrasp, zhangcadgrasp}, severely compromising inference efficiency.

To address these data and algorithmic bottlenecks, we introduce a holistic paradigm comprising a large-scale cluttered scene  benchmark and a novel end-to-end generation framework. To bypass the prohibitive computational cost of scene-level optimization, our OmniDex benchmark employs a scalable seed-and-filter strategy. We first establish a reusable library of free-space seed grasps for single objects, then filter them through collision constraints in cluttered scenes within gravity-stabilized simulations. Leveraging high-quality 3D object assets and high-fidelity rendering, this pipeline yields multi-modal RGB-D observations from multiple fixed camera views, aligned with scene-aware grasp ground truths. Ultimately, our OmniDex benchmark delivers unprecedented scale: over 2.6 million scenes and 0.4B scene-specific grasp pose labels, covering diverse objects, supporting bases, and realistic cluttered layouts. We hope this benchmark will spur a new wave of research in cluttered scenes dexterous grasping. Empowered by this benchmark, OmniDex model fundamentally avoids the pitfalls of non-end-to-end intermediate representations and sluggish test-time fine-tuning. Instead, we internalize physical dynamics directly into training. By coupling a Soft Winner-Takes-All learning to capture grasp multimodality with a human-inspired physics-constrained loss—which enforces direct hand-object interaction—we effectively bridge the critical last-millimeter precision gap to ensure close and stable grasping. During inference, the optimal grasp pose is rapidly identified via a physical ranking mechanism, achieving both exceptional fidelity and real-time efficiency.

\begin{table}[!t]
    \centering
    \caption{Quantitative comparison of simulation-based dexterous grasping benchmarks. Our dataset features the large scale of scenes and objects, emphasizing diverse layouts in clutter.}
    \label{table1}
    \resizebox{0.7\textwidth}{!}{
    \begin{tabular}{l c c c l c}
        \toprule
        \textbf{Method} & \textbf{\# Scenes} & \textbf{Setting} & \textbf{\# Objects} & \textbf{Workspace} & \textbf{Modality} \\
        \midrule
        Dex1B \cite{yedex1b} & - & Single & 6K & - & Depth \\
        AffordDexGrasp \cite{wei2025afforddexgrasp} & 2K & Single & 1K & Table & RGB-D \\
        DDGC \cite{lundell2021ddgc} & 0.4K & Cluttered & 0.3K & Table & RGB-D \\
        DexGraspNet 2.0 \cite{zhang2024dexgraspnet} & 8K & Cluttered & 1K & Table & Depth \\
        ClutterDexGrasp \cite{chen2025clutterdexgrasp} & 1K & Cluttered & 2K & Table & Depth \\
        \midrule
        \rowcolor{gray!15} \textbf{Ours} & \textbf{2.6M} & \textbf{Cluttered} & \textbf{6K} & \textbf{Table, Box, Shelf} & \textbf{RGB-D} \\
        \bottomrule
    \end{tabular}
    }
\end{table}

\noindent\textbullet\ We introduce a massive cluttered-scene OmniDex benchmark via a scalable seed-and-filter paradigm, yielding an unprecedented dataset of over 2.6 million realistic scenes and 0.4B scene-specific validated grasp ground truths, featuring diverse supporting bases and rich multi-modal observations.

\noindent\textbullet\ We propose OmniDex model, a novel end-to-end grasp generation framework that couples Soft Winner-Takes-All learning with human-inspired physical constraints to capture grasp multimodality and bridge the last-millimeter precision gap, with physics-driven ranking for robust inference.

\noindent\textbullet\ Experimental results show that OmniDex achieves state-of-the-art performance on the OmniDex benchmark. It generalizes well across scene subsets, supporting bases, input views, and unseen objects, demonstrating robustness in realistic cluttered scenes.

\section{Related Works}
\label{sec2}
\subsection{Dexterous Grasping Datasets and Benchmarks}
\label{sec2.1}
Real-world dexterous grasping datasets largely rely on human teleoperation \cite{liu2024realdex, wang2025dexh2r, xudexumi, hang2024dexfuncgrasp}. While capturing human-like priors, this labor-intensive paradigm limits dataset scale and object diversity. Consequently, the community shifted to simulation-based synthesis for isolated objects. Pioneering works \cite{wang2023dexgraspnet, zhong2025dexgrasp, chendexonomy, wan2023unidexgrasp++} leveraged optimization and physical priors to generate massive single-object grasps, culminating in billion-scale datasets like Dex1B \cite{yedex1b}. Despite this impressive scale, these single-object benchmarks fail to model the spatial dynamics of real-world clutter. While recent efforts have introduced cluttered-scene benchmarks \cite{zhang2024dexgraspnet, chen2025clutterdexgrasp} to bridge this reality gap, they remain bottlenecked by costly per-scene optimization. In stark contrast, our efficient seed-and-filter strategy entirely bypasses this sluggish optimization, yielding an unprecedented 2.6 million complex scenes and 0.4B physically validated ground truth, thereby establishing a new benchmark for diversity and scale in cluttered grasping.

\subsection{Dexterous Grasp Generation Methods}
\label{sec2.2}

Learning-based generative models promise real-time dexterous grasp generation, yet mapping visual inputs to high-DoF kinematics is fundamentally plagued by grasp multimodality. To capture this one-to-many manifold, data-driven methods typically rely on heuristic proxies—such as contact maps \cite{xu2023unidexgrasp, zhangcadgrasp} or affordance fields \cite{zhang2024dexgraspnet, qin2023dexpoint}, making the pipeline non-end-to-end. Furthermore, these models struggle with the last-millimeter precision, where minute deviations cause catastrophic contact failures. To mitigate this, recent frameworks \cite{wei2025afforddexgrasp,he2025dexvlg} use computationally heavy test-time optimization \cite{chen2025bodex}, compromising real-time efficiency. While some attempts internalize physical priors during training to bypass this latency \cite{zhong2025dexgrasp}, their geometric penalties remain insufficient for dense clutters. In stark contrast, our framework resolves these bottlenecks: we capture the multimodal distribution via a Soft Winner-Takes-All objective, and directly bridge the precision gap using a human-inspired physical constraint during training, coupled with a physical ranking during inference.

\section{Benchmark}
\label{sec3}
\subsection{Overview of Data Generation}
\label{sec3.1}
Our data generation pipeline targets realistic cluttered scenes while producing two tightly coupled resources: (i) state observation representations; (ii) instruction-conditioned dexterous hand grasp pose labels. A central difficulty arises from the high DoF of dexterous hands: we must generate grasps that are not only physically feasible but also executable in cluttered scenes. Optimization-based methods have proven effective for dexterous hand grasp generation for a single object and can be applied in tabletop settings. However, the tabletop setting is inherently instance-specific, as each run is anchored to a particular object–table configuration. Moreover, its computational cost is prohibitive (e.g., optimizing 500 initializations for one object takes ${\sim}$ 15 minutes on an A800), making per-object, per-scene optimization infeasible for large-scale cluttered scene benchmark construction.

\begin{figure*}[!tbp] 
\begin{center}
\includegraphics[width=0.65\linewidth]{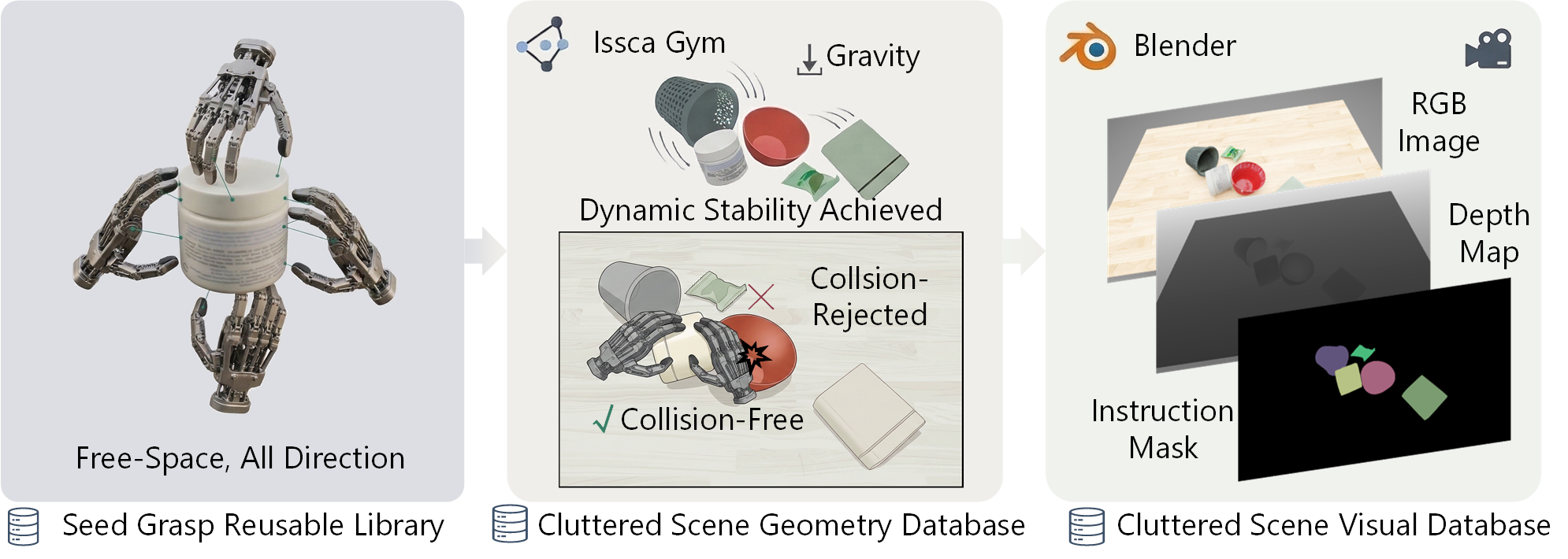}  
\end{center}
\caption{\textbf{Seed-and-filter generation pipeline}. Our two-stage framework generates scene-valid grasps by: (1) constructing a reusable seed grasp library in free space, and (2) simulating cluttered scenes in Isaac Gym for collision filtering, followed by multi-modal rendering in Blender.}
\label{fig1}
\end{figure*}

As shown in Fig. \ref{fig1}, to enable scalable generation, we adopt a two-stage seed-and-filter strategy. First, we run optimization for each object in free space to produce a reusable library of seed grasp poses sampled from diverse directions. Next, we simulate cluttered scenes in Isaac Gym by dropping objects onto supporting bases. Once dynamic stability is achieved, we transform free-space seed grasp poses into scene-specific candidates and retain those that satisfy a conservative direct-retrieval criterion. Specifically, accepted candidates must establish target contact while avoiding support surfaces and clearly disruptive interactions with neighboring objects. This protocol is designed to isolate direct target extraction in cluttered scene, rather than rearrangement-mediated retrieval. To simulate the visual scene observations available to learning agents, we replay the cluttered scene geometry in Blender to build a visual database, including RGB images, depth maps, semantic segmentation, and camera parameters. The segmentation taxonomy is defined within each scene and used to generate target-object masks as visual instructions for grasping. In summary, the visual database provides diverse model inputs, while the geometry database provides scene-valid grasp ground truths for both model supervision and evaluation. Qualitative visualizations are in Appendix \ref{AppF.1}.

\subsection{3D Assets}
\label{sec3.2}

\begin{figure}[!t] 
\begin{center}
\includegraphics[width=0.95\linewidth]{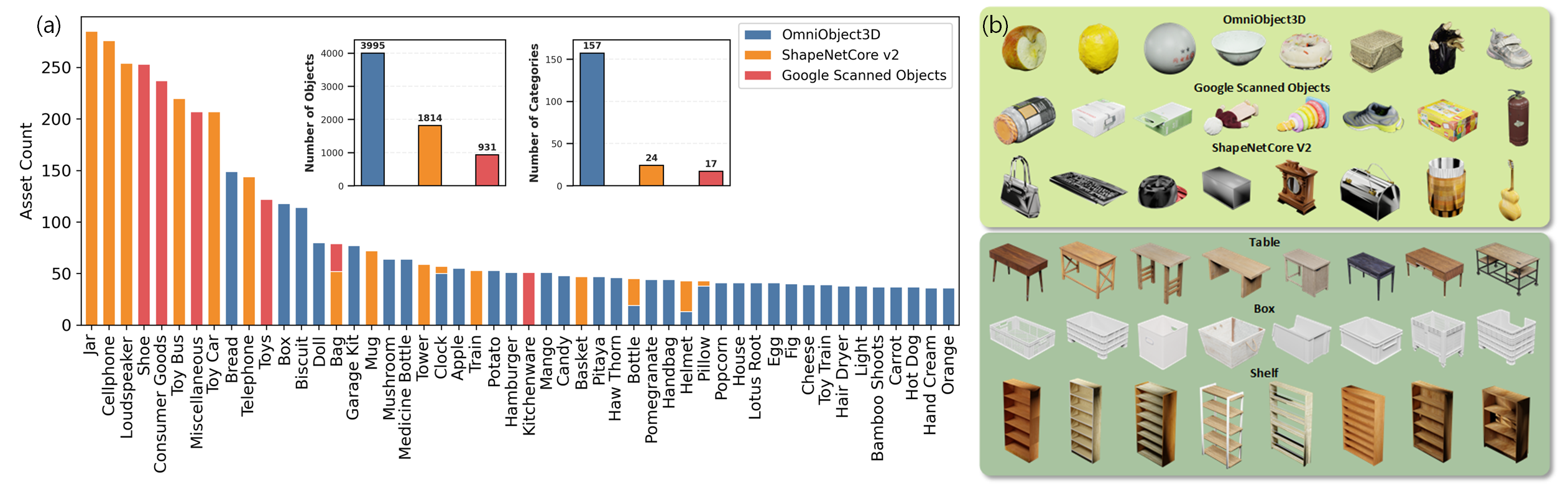}  
\end{center}
\caption{\textbf{Dataset statistics and visualization of 3D assets}. (a): Instance counts and category-wise distributions of the curated graspable objects, highlighting the top 50 semantic classes. (b): Visual examples of the assets, which range from diverse grasped objects (medicine bottles, boxes, apples) to realistic supporting bases (tables, open-top boxes, shelves).}
\label{fig2}
\end{figure}

\noindent \textbf{Grasped Objects.} We curate graspable object assets from OmniObject3D \cite{wu2023omniobject3d}, ShapeNetCore v2 \cite{chang2015shapenet}, and Google Scanned Objects \cite{downs2022google} datasets. The composition of the curated assets, including instance counts and category-wise distributions, is detailed in Fig. \ref{fig2} (a). Our selection follows a staged protocol: we first use their semantic annotations to exclude categories incompatible with dexterous hand grasping, then enforce inter-class balancing to avoid long-tailed bias, and finally remove instances with defective meshes or incomplete textures. Each asset is processed into two scale-aligned representations: a physics-ready model generated via manifold–normalization–decomposition \cite{wang2023dexgraspnet} and a high-fidelity visual model, both sharing an identical normalization to ensure spatial consistency. In total, we provide a database of over 6,000 high-quality objects, spanning diverse categories such as cellphones, toys, and shoes. All selected instances are manually inspected to ensure they are suitable for the dexterous grasping task.

\noindent \textbf{Supporting Base.} As illustrated in Fig. \ref{fig2} (b), our OmniDex benchmark incorporates tables, open-top boxes, and shelves to diversify object placement in cluttered scenes. Specifically, our asset library comprises 244 tables from ABO \cite{collins2022abo}, 79 open-top boxes sourced from both ABO and Isaac Lab \cite{mittal2025isaac}, and 21 shelves from ShapeNetCore v2 \cite{chang2015shapenet}.

\subsection{Scene-Specific Ground-Truth Grasp Pose Generation}
\label{sec3.3}
Following the seed-and-filter strategy in Sec. \ref{sec3.1}, we first optimize 4,000 candidates per object using the force-closure estimator \cite{wang2023dexgraspnet}, retaining those that pass a simulated lift test as seed grasp poses. To construct cluttered scenes, a random number of ${N}$ objects are dropped from stochastic heights, spatially confined within the horizontal bounds of the supporting base. We adopt subset-specific scene generation rules to induce distinct clutter regimes: random tabletop clutter for $\mathcal{D}_{\text{std}}$ and $\mathcal{D}_{\text{mix}}$, dense confined stacking for $\mathcal{D}_{\text{box}}$, spaced identical-object layouts for $\mathcal{D}_{\text{grid}}$, and layer-wise arrangements for $\mathcal{D}_{\text{shelf}}$. Please see Appendix \ref{AppA.2} for more details.

Simulation terminates once all objects reach stable resting conditions or a fixed rollout cutoff is reached. The resulting settled scene yields stable object poses ${{\bf{o}} = \left( {{{\bf{t}}_{\rm{obj}}},{{\bf{R}}_{\rm{obj}}}} \right)}$, where ${{{\bf{t}}_{\rm{obj}}} \in {{ \mathbb{R} }^3}}$ and ${{{\bf{R}}_{\rm{obj}}} \in SO\left( 3 \right)}$ denote the object translation and orientation, respectively.  We then apply the object transformations to the selected seed grasps to derive scene-specific poses ${{\bf{g}} = \left( {{{\bf{t}}_{\rm{dex}}},{{\bf{R}}_{\rm{dex}}},\theta } \right)}$. Here, ${{{\bf{t}}_{\rm{dex}}} \in {{ \mathbb{R} }^3}}$ and ${{{\bf{R}}_{\rm{dex}}} \in SO\left( 3 \right)}$ denote the wrist translation and orientation, respectively, while ${\theta }$ represents joint angles. The complete scene configuration is thus denoted as ${{\bf{s}} = \left( {{\bf{o}},{\bf{g}}} \right)}$. To mitigate artifacts from physical-visual mesh discrepancies and incomplete simulation convergence, we perform a sanity check by replaying each scene with visual models. Please see Appendix \ref{AppA.1} for more details.

\subsection{Diversity and Protocol}
\label{sec3.4}

\textbf{Modal Diversity.} As discussed in Sec. \ref{sec1}, geometry-only input is inadequate for instruction-guided grasping in cluttered scenes. We therefore provide aligned RGB, depth, instruction masks, and camera parameters under a subset-aware multi-view protocol. We use one ego-centric view and four auxiliary views by default, while adapting the camera layout to each subset: tabletop scenes use standard corner views, box scenes tighten the rig to counter boundary occlusion, and shelf scenes adopt layer-aligned frontal views to maintain visibility under severe vertical blockage. This makes perception a first-class benchmark component. Please see Appendix~\ref{AppA.2} for details.

\begin{table*}[t]
\centering
\caption{\textbf{Benchmark diversity and evaluation protocol across OmniDex subsets.} Each subset is organized with a large training split and two decoupled test tracks to evaluate generalization.}
\label{Table2}
\resizebox{0.92\textwidth}{!}{
\begin{tabular}{lccccccccc}
\toprule
\multirow{2}{*}{Subset} & \multirow{2}{*}{Obj. Div.} & \multirow{2}{*}{Scene Obj. Comp.} & \multirow{2}{*}{Sup. Base Type} & \multirow{2}{*}{Sup. Base Div.} & \multirow{2}{*}{Density} & \multirow{2}{*}{Scene Num.} & \multicolumn{3}{c}{Generalization Split} \\
\cmidrule(lr){8-10}
 &  &  &  &  &  &  & Train & Seen & Unseen \\
\midrule2
${{{\cal D}_{{\rm{std}}}}}$   & \ding{51} & Multi-category  & Table & \ding{55} & 0.1 / 0.2 / 0.3 & 1.14M & 919K & 62K & 163K \\
${{{\cal D}_{{\rm{mix}}}}}$   & \ding{51} & Multi-category  & Table & \ding{51} & 0.1 / 0.2 / 0.3 & 0.94M & 706K & 80K & 152K \\
${{{\cal D}_{{\rm{box}}}}}$   & \ding{51} & Multi-category  & Box   & \ding{51} & 0.9             & 0.14M & 112K & 12K & 16K \\
${{{\cal D}_{{\rm{grid}}}}}$  & \ding{51} & Single-category & Table & \ding{55} & -               & 0.28M & 246K & 11K & 24K \\
${{{\cal D}_{{\rm{shelf}}}}}$ & \ding{51} & Multi-category  & Shelf & \ding{51} & -               & 0.14M & 107K & 11K & 28K \\
\bottomrule
\end{tabular}
}
\end{table*}

\textbf{Task Diversity.} To align with the core objective of dexterous grasping, our OmniDex benchmark introduces diversity across: grasped object and supporting base. These factors directly dictate grasp feasibility and perceptual difficulty. As illustrated in Table \ref{Table2}, following prior works \cite{yedex1b, he2025dexvlg}, we provide ${{{\cal D}_{{\rm{std}}}}}$ with 1.14M scenes where the supporting base remains fixed, varying only the target object. This design isolates the evaluation of policy transferability across diverse geometries and appearances. To evaluate grasping in more unstructured environments, we include ${{{\cal D}_{{\rm{mix}}}}}$ with 0.94M scenes where both the target objects and the supporting bases vary simultaneously.

\textbf{Real Scene Diversity.} To approximate real-scene applications, we introduce three specialized ${{{\cal D}_{{\rm{box}}}}}$, ${{{\cal D}_{{\rm{grid}}}}}$, and ${{{\cal D}_{{\rm{shelf}}}}}$. As illustrated in Fig. \ref{fig0}, ${{{\cal D}_{{\rm{box}}}}}$ models parcel sorting, where objects are densely packed both vertically and horizontally within confined containers, mimicking carton or express package handling. ${{{\cal D}_{{\rm{grid}}}}}$ simulates warehouse sorting environments featuring multiple identical objects uniformly distributed on a table. To ensure stable post-drop configurations in this subset, we restrict our asset pool to objects capable of remaining upright after falling (e.g., bottles). Finally, ${{{\cal D}_{{\rm{shelf}}}}}$ targets SKU-level retail scenarios where products are systematically arranged on shelves; for this, we specifically curate retail-style items (e.g., noodle cups) from complete object dataset.

\textbf{Benchmark Protocol.} As summarized in Table \ref{Table2}, each subset is partitioned into a large training split and two test tracks: \emph{Seen Objects in Unseen Layouts} and \emph{Unseen Objects in Unseen Layouts}. The former isolates spatial reasoning under novel clutter configurations while controlling for object familiarity, whereas the latter probes zero-shot transfer to unseen geometries under equally novel layouts. This protocol cleanly separates layout novelty from object novelty, yielding a benchmark that measures robust scene-level generalization in cluttered dexterous grasping.

\subsection{Metrics}
\label{sec5.2}
Aligned with our OmniDex benchmark protocol, we evaluate model performance along two dimensions: physical grasp stability and protocol-consistent scene interaction under direct retrieval. Following \cite{wang2023dexgraspnet}, a grasp is deemed physically valid if the target object securely withstands applied gravity with the supporting surface removed. We quantify this via three Success Rate ($\text{SR}$) variants based on the tested gravitational directions: the actual scene gravity ($\text{SR}_{\text{scene}}$, our primary metric), all six canonical directions ($\text{SR}_{\text{any}}$), or at least one canonical direction ($\text{SR}_{\text{one}}$). To measure consistency with our selection protocol, we further report the Collision-Free Rate ($\text{CFR}$), defined as the percentage of generated grasps that avoid support-surface collisions and non-target contacts under the simulator.

\section{Methodology}
\label{sec4}

\subsection{Overview}
\label{sec4.1}
For clarity, we present the ego-centric single-view formulation here. We aim to generate a valid grasp pose ${{\bf{\hat g}}}$ for a target object in a cluttered scene, given an RGB image ${{\bf{I}} \in {{ \mathbb{R} }^{H \times W \times 3}}}$, a depth map ${{\bf{D}} \in {{ \mathbb{R} }^{H \times W \times 1}}}$, camera parameters ${\left\{ {{\bf{K}},{\bf{T}}} \right\}}$, the table height ${\bf{H}}$, and an instruction mask ${{\bf{M}} \in {\left\{ {0,1} \right\}^{H \times W}}}$. Here, the instruction mask explicitly specifies the target object within the clutter. ${{\bf{K}}}$ and ${{\bf{T}}}$ denote the intrinsic and extrinsic camera matrices, respectively, while ${H}$ and ${W}$ denote the image height and width. Our method is formulated as training a diffusion-based neural network to generate a high-DoF dexterous grasp pose conditioned on scene appearance, geometry, and target identity: $\mathbf{\hat g} = \Theta \left( \mathbf{I}, \mathbf{D}, \mathbf{M}, \mathbf{K}, \mathbf{T}, \mathbf{H} \right)$. The multi-view extension is provided in Appendix~\ref{AppC.2}.

\begin{figure*}[!tb] 
\begin{center}
\includegraphics[width=0.95\linewidth]{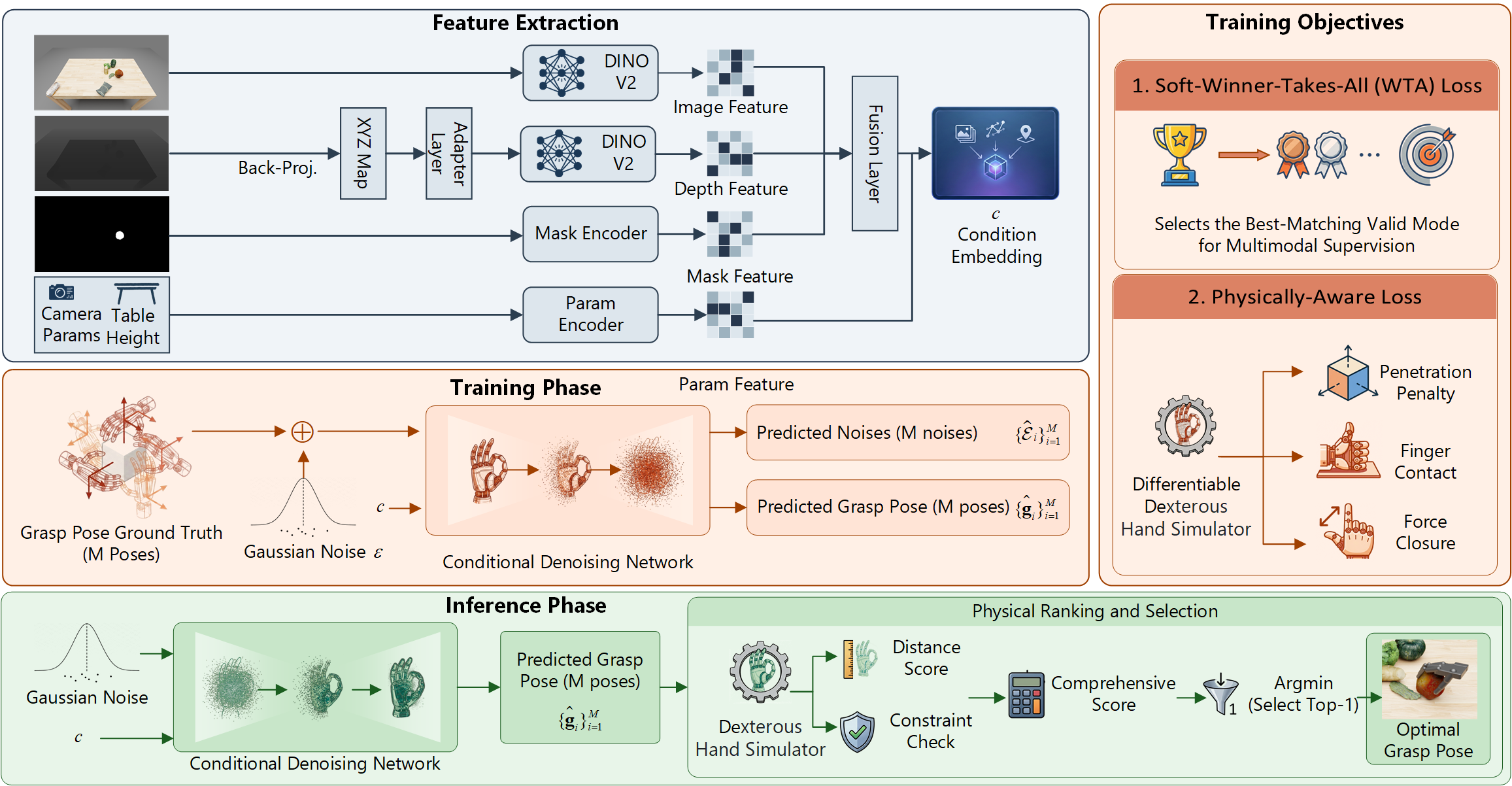}  
\end{center}
\caption{\textbf{Architecture of the OmniDex Model}. Condition embeddings are transformed from RGB-D inputs, instruction masks, and scene parameters. During training, a denoising network adds identical noise to $M$ grasp poses, optimized jointly via a Soft Winner-Takes-All loss (for multimodal distributions) and a physics-aware loss (for stable contact). During inference, the network generates $M$ candidates, followed by a physics-based ranking module to select the optimal grasp.}
\label{fig4}
\end{figure*}

The overview of the OmniDex model is illustrated in Fig. \ref{fig4}. In the feature encoder module, we extract features from the camera image, depth map, and instruction mask, and aggregate them into a unified perception feature for the cluttered scene. It is subsequently combined with the scene parameter features to yield the condition embedding (Sec. \ref{sec4.2}). In the training phase, we apply a Soft Winner-Takes-All (WTA) loss to address the inherently multimodal distribution of grasp poses, alongside a novel physics-constrained loss designed to enforce close and stable picking (Sec. \ref{sec4.3}). In the inference phase, the model generates multiple grasp pose candidates, from which the optimal grasp is identified via a physical ranking and selection (Sec. \ref{sec4.4}).

\subsection{Multimodal Scene-Aware Condition Embedding}
\label{sec4.2}
To generate accurate and stable dexterous grasps in cluttered scenes, the model requires a comprehensive understanding of both the semantic and geometric structure of the workspace. Under the ego-centric single-view setting described above, we construct a multimodal, scene-aware condition embedding by fusing the RGB image, depth map, instruction mask, and scene parameters into a unified tokenized representation.

Given the ego-centric RGB image, we use a pre-trained DINOv2 \cite{oquab2024dinov2} backbone to extract patch-level semantic features ${{{\bf{F}}_{\rm{img}}} \in {{ \mathbb{R} }^{N \times d}}}$, where ${N}$ is the number of patches and ${d}$ is the embedding dimension. We derive geometric features ${{{\bf{F}}_{\rm{depth}}} \in {{ \mathbb{R} }^{N \times d}}}$ from the depth map via a lightweight adapter, and encode the instruction mask into mask features ${{{\bf{F}}_{\rm{mask}}} \in {{ \mathbb{R} }^{N \times d}}}$. We then fuse semantic, geometric, and mask features to obtain a scene-aware representation ${{{\bf{F}}_{\rm{scene}}} \in {{ \mathbb{R} }^{N \times d}}}$.

To anchor grasp generation in a consistent global frame, we additionally encode the camera parameters and table height into a parameter token ${{{\bf{f}}_{{\rm{para}}}} \in {{ \mathbb{R} }^{1 \times d}}}$. Appending ${{{\bf{f}}_{{\rm{para}}}}}$ to ${{{\bf{F}}_{{\rm{scene}}}}}$ yields the final condition embedding ${{\bf{c}} \in {{ \mathbb{R} }^{\left( {N + 1} \right) \times d}}}$. Implementation details are deferred to Appendix~\ref{AppC.1}.

\subsection{Physically-Constrained Denoising for Multimodal Grasp Generation}
\label{sec4.3}

Cluttered-scene grasping is intrinsically multimodal: under the same scene observation, a target object may admit multiple valid grasp modes. Deterministic supervision tends to average across these modes, often producing implausible intermediate poses. Inspired by winner-takes-all learning \cite{rupprecht2017learning}, we train the diffusion model with a soft WTA-style objective that preserves this one-to-many structure directly in grasping space. Specifically, we sample a set of ${M}$ valid grasp poses for a target object and inject the same Gaussian noise ${\varepsilon}$ into all ${M}$ poses at diffusion step ${t}$. The network is then trained to explain the corrupted sample through the most compatible grasp mode rather than their average. Here, ${M}$ serves as a finite hypothesis budget for approximating the dominant modes of the conditional grasp distribution. The resulting Soft-WTA loss is
\begin{equation}
\label{eq4}
\mathcal{L}_{\text{WTA}} = \mathbb{E}_{t,\varepsilon} \Big[ -\tau \log \sum_{i = 1}^M \exp \big( - \|\varepsilon - \hat{\varepsilon}_i\|^2 / \tau \big) \Big],
\end{equation}
where $\hat{\varepsilon}_i$ is the predicted noise for the $i$-th grasp hypothesis and $\tau$ is a temperature parameter. This objective allows the model to represent diverse grasp modes without introducing handcrafted intermediate representations.

Purely data-driven denoising, however, is often insufficient for dexterous grasping, where millimeter-scale errors can break contact or induce unstable force closure. We therefore augment training with a physics-constrained objective defined on a differentiable dexterous hand simulator:
\begin{equation}
\label{eq_total}
\mathcal{L}_{\text{total}} = \lambda_{\text{WTA}} \mathcal{L}_{\text{WTA}} + \lambda_{\text{phys}} (\mathcal{L}_{\text{thumb}} + \mathcal{L}_{\text{opp}} + \mathcal{L}_{\text{SDF}} + \mathcal{L}_{\text{closure}}).
\end{equation}
Here, $\mathcal{L}_{\text{thumb}}$ and $\mathcal{L}_{\text{opp}}$ encourage stable contact on the thumb and opposing fingers, $\mathcal{L}_{\text{closure}}$ promotes an antipodal grasp configuration, and $\mathcal{L}_{\text{SDF}}$ penalizes hand-object interpenetration. Full definitions are provided in Appendix~\ref{AppC.3}.

\subsection{Physics-Driven Ranking and Selection}
\label{sec4.4}
In the inference stage, a single sampled pose may exhibit geometric misalignment due to the inherent stochasticity of generative models. To enable robust execution, we introduce a parallel generation and physics-driven ranking paradigm. Specifically, given the same condition embedding $\mathbf{c}$, we simultaneously sample $M$ independent Gaussian noises to decode candidate grasp poses $\{ \hat{\mathbf{g}}_i \}_{i=1}^M$. These candidates are evaluated in the dexterous hand simulator to compute a comprehensive physical score $S(\hat{\mathbf{g}})$, assessing kinematic closure while strictly penalizing physical violations:

\begin{equation}
\label{eq8}
S(\hat{\mathbf{g}}) = S_{\text{kin}}(\hat{\mathbf{g}}) - R_{\text{con}}(\hat{\mathbf{g}}) + \Omega_{\text{phy}}(\hat{\mathbf{g}})
\end{equation}

The primary metric $S_{\text{kin}}(\hat{\mathbf{g}})$ evaluates force closure stability via finger-to-object distances as Sec. \ref{sec4.3}. The remaining terms act as essential constraint checks: the contact reward $R_{\text{con}}(\hat{\mathbf{g}})$ encourages multi-finger engagement by rewarding additional fingertip contacts, while the safety penalty $\Omega_{\text{phy}}$ linearly penalizes slight interpenetrations and heavily rejects severe collisions or missing thumb-opposition. Ultimately, we filter out physically invalid modes to output the optimal grasp pose.

\section{Experiment}
\label{sec5}

\subsection{Implementation Details}
\label{sec5.1}
We instantiate our framework on the Shadow Dexterous Hand \cite{shadow_hand}, leveraging its high-dimensional action space (24 DoF) to rigorously validate our model's generative capacity. All subsets of our proposed benchmark adopt an 8/2 split for the train and test sets. The test sets are decoupled into two distinct tracks: Seen objects in unseen clustered scene; unseen objects (see Appendix \ref{AppB}). Please refer to Appendix \ref{AppD} for more details.

\subsection{Main Results}
\label{sec5.3}

As shown in Table \ref{table2.5}, our method establishes the state-of-the-art in cluttered-scene dexterous grasping. All baselines in Table~\ref{table2.5} are retrained on the same OmniDex training split, rather than evaluated zero-shot. On $\mathcal{D}_{\text{std}}$ with ego-view input, OmniDex achieves 55.80 SR$_{\text{scene}}$ and 84.12 CFR, surpassing DexGraspVLA by 14.81 and 2.68 points, respectively. This advantage also holds for unseen objects, where OmniDex attains 59.74 SR$_{\text{scene}}$, demonstrating generalization beyond the training set. Across subsets, our method remains effective across scene layouts and support types, and reaches particularly strong performance on $\mathcal{D}_{\text{grid}}$, indicating reliable use of structured spatial cues. Overall, these results verify that the OmniDex model delivers superior accuracy, robustness, and cross-setting generalization for realistic cluttered-scene grasping. Qualitative results are shown in Appendix~\ref{AppF.2}.

\begin{table}[htbp]
  \centering
  \caption{\textbf{Main Results on the OmniDex Benchmark}. We evaluate our method against baselines across diverse scene subsets and compare with other methods on $\mathcal{D}_{\text{std}}$. * denotes the use of the finger-force trick \cite{wang2023dexgraspnet}}
  \label{table2.5}
  \resizebox{1.0\textwidth}{!}{%
  \begin{tabular}{llc rcrcccc rcrcccc} 
    \toprule
    \multirow{2}{*}{Method} & \multirow{2}{*}{Subset} & \multirow{2}{*}{\shortstack{Input\\View}} & \multicolumn{7}{c}{Seen Objects} & \multicolumn{7}{c}{Unseen Objects} \\
    \cmidrule(lr){4-10} \cmidrule(lr){11-17}
    & & & SR$_{\text{scene}}$ & SR$_{\text{scene}}^*$ & SR$_{\text{any}}$ & SR$_{\text{any}}^*$ & SR$_{\text{one}}$ & SR$_{\text{one}}^*$ & CFR & SR$_{\text{scene}}$ & SR$_{\text{scene}}^*$ & SR$_{\text{any}}$ & SR$_{\text{any}}^*$ & SR$_{\text{one}}$ & SR$_{\text{one}}^*$ & CFR \\
    \midrule
    \multirow{10}{*}{Ours} 
    & $\mathcal{D}_{\text{std}}$   & Ego   & 33.60  & 55.80  & 18.87 & 36.79 & 65.15 & 79.35  & 84.12 & 40.15 & 59.74 & 27.18 & 40.77 & 65.54 & 81.77 & 84.61 \\
    & $\mathcal{D}_{\text{mix}}$   & Ego   & 32.05  & 52.21  & 17.12 & 34.04 & 58.51 & 74.52  & 73.12 & 37.90 & 55.10 & 24.13 & 35.19 & 62.41 & 78.02 & 83.22 \\
    & $\mathcal{D}_{\text{box}}$   & Ego   & 32.34  & 35.83  & 17.06 & 19.79 & 55.15 & 56.52  & 99.12 & 33.48 & 36.50 & 17.80 & 20.17 & 56.77 & 58.11 & 95.72 \\
    & $\mathcal{D}_{\text{grid}}$  & Ego   & 55.01  & 79.60  & 33.68 & 55.02 & 91.91 & 93.70  & 95.46 & 43.11 & 68.05 & 33.97 & 52.67 & 87.50 & 92.09 & 94.86 \\
    & $\mathcal{D}_{\text{shelf}}$ & Ego   & 29.49  & 39.46  & 18.31 & 25.25 & 55.17 & 60.57  & 81.95 & 14.96 & 24.08 & 11.58 & 15.08 & 34.26 & 37.89 & 63.57 \\
    \cmidrule{2-17} 
    & $\mathcal{D}_{\text{std}}$   & Multi & 37.35 & 60.65 & 22.18 & 41.18 & 68.01 & 82.60  & 85.11 & 45.05 & 64.08 & 30.79 & 43.76 & 68.48 & 84.50 & 87.24 \\
    & $\mathcal{D}_{\text{mix}}$   & Multi & 35.41 & 56.02 & 20.54 & 38.31 & 62.47 & 78.06  & 75.31 & 42.56 & 60.34 & 28.36 & 40.12 & 65.83 & 82.61 & 85.04 \\
    & $\mathcal{D}_{\text{box}}$   & Multi & 27.89 & 28.52 & 19.11 & 19.53 & 41.20 & 42.06  & 97.91 & 31.14 & 31.71 & 21.17 & 21.79 & 43.33 & 44.30 & 97.64 \\
    & $\mathcal{D}_{\text{grid}}$  & Multi & 60.37 & 84.28 & 38.44 & 58.95 & 95.14 & 96.52  & 96.49 & 46.68 & 73.14 & 37.42 & 56.82 & 91.21 & 95.31 & 95.71 \\
    & $\mathcal{D}_{\text{shelf}}$ & Multi & 34.06 & 44.63 & 22.27 & 29.61 & 59.88 & 64.31  & 83.42 & 18.21 & 28.86 & 15.24 & 19.37 & 39.54 & 42.21 & 65.29 \\
    \midrule
    DexGraspVLA \cite{zhong2026dexgraspvla} & $\mathcal{D}_{\text{std}}$     & Ego & 24.30    & 40.99  & 11.87 & 24.42  & 53.83  & 67.05     & 81.44 & 39.42   & 54.97 & 29.95 & 41.02& 54.89  & 71.55 & 42.86 \\
    DexGraspNet 2.0 \cite{zhang2024dexgraspnet} & $\mathcal{D}_{\text{std}}$ & Ego & 28.03    & 43.64  & 13.16 & 25.59  & 56.04  & 68.47     & 81.06 & 32.40   & 47.60 & 18.83 & 27.39 & 58.77 & 73.73 & 85.15 \\
    Grasp-as-You-Say \cite{wei2024grasp} & $\mathcal{D}_{\text{std}}$        & Ego & 0.24     & 1.68   & 0.09  & 0.15   & 20.91  & 27.45     & 38.57 & 0.44    & 2.55 & 0.10 & 0.42 & 19.46    & 25.57 & 64.25 \\
    \bottomrule
  \end{tabular}%
  }
\end{table}

\subsection{Ablation Studies}
\label{sec5.4}

We evaluate our procedures on a representative mini-split curated from ${{{\cal D}_{{\rm{std}}}}}$ (see Appendix \ref{AppB}). By default, baselines operate on a single ego-centric view. We use $\text{SR}_{\text{scene}}$ as the principal metric to quantify grasp stability. More ablation studies can be seen in Appendix \ref{AppE}.

\textbf{Effect of Benchmark Scale}. To justify the massive scale of our proposed OmniDex benchmark, we train our model on incrementally larger fractions of the training set, and evaluate it on the full test set. As Table \ref{table3} shows, performance scales strictly with data volume. Scaling explicitly enhances the model's capacity to reason about complex inter-object physical interactions in cluttered layouts; notably, CFR surges from 81.54\% (20\% subset) to 83.66\% (100\% subset). This significant improvement in stable grasping validates the necessity of our large-scale data generation pipeline.

\begin{table}[htbp]
\centering
\caption{Quantitative results when trained on incrementally larger fractions of training dataset.}
\label{table3}
\resizebox{0.4\linewidth}{!}{
\begin{tabular}{lccccc}
\toprule
Metrics & 20\% & 40\% & 60\% & 80\% & 100\% \\
\midrule
$\text{SR}_{\text{scene}}$ & 40.50 & 41.04 & 43.84 & 44.32 & 44.41 \\
CFR & 81.54 & 82.64 & 82.13 & 84.15 & 83.66 \\
\bottomrule
\end{tabular}
}
\end{table}

\textbf{Effect of Input Modalities}. We analyze the role of various inputs as shown in Table \ref{table4}. Depth provides the foundational shape cognition. RGB supplies scene semantics and, more crucially, localizes the target object within the cluster. Camera parameters ensure accurate geometric perception. Multi-view inputs yield significant performance gains by enabling comprehensive scene understanding. Table height serves as a vital spatial anchor; since the coordinate origin lies at the table center, it aligns the depth-derived geometry with grasp ground truths to explicitly dictate precise grasp locations.

\begin{table}[htbp]
\centering
\caption{Quantitative ablation on various input conditions. The highlighted row represents baseline.}
\label{table4}
\resizebox{0.6\linewidth}{!}{
\begin{tabular}{ccccccc}
\toprule
RGB & Depth & Cam. Para. & Table Height & Multi-view & $\text{SR}_{\text{scene}}$ & CFR \\
\midrule
\checkmark &            & \checkmark & \checkmark &            & 31.89 & 79.97 \\
           & \checkmark & \checkmark & \checkmark &            & 36.80 & 82.81 \\
\checkmark & \checkmark &            & \checkmark &            & 43.68 & 84.09 \\
\checkmark & \checkmark & \checkmark &            &            & 40.64 & 84.04 \\
 \checkmark & \checkmark &\checkmark & \checkmark & \checkmark & 49.87 & 82.11  \\
\rowcolor{gray!15}\checkmark & \checkmark & \checkmark & \checkmark &            & 44.41 & 83.66 \\
\bottomrule
\end{tabular}
}
\end{table}

\textbf{Effect of Model Components}. Table \ref{table5} details our architectural ablations. Standard denoising loss traps generation in an averaging effect across grasp hypotheses. Conversely, $\mathcal{L}_{\rm{WTA}}$ resolves this multi-modal manifold issue, substantially boosting $\text{SR}_{\text{scene}}$. Training-phase physical constraints are vital for overcoming the critical last-millimeter contact challenge. At inference, our physics-driven ranking drastically outperforms naive random sampling, elevating $\text{SR}_{\text{scene}}$ from 30\% to 40\%. Test-time optimization yields limited gains, bottlenecked by inherent instability and depth-backprojection occlusions (see details in Appendix \ref{AppE}).

\begin{table}[htbp]
\centering
\caption{Component-wise ablation study evaluated on both training and inference phases. TTP denotes Test-Time Optimization.}
\label{table5}
\resizebox{0.7\linewidth}{!}{
\begin{tabular}{cccccccccc}
\toprule
\multicolumn{5}{c}{Training Phase} & \multicolumn{3}{c}{Inference Phase} & \multicolumn{2}{c}{Metrics} \\
\cmidrule(lr){1-5} \cmidrule(lr){6-8} \cmidrule(lr){9-10}
$\mathcal{L}_{\text{stand}}$ & $\mathcal{L}_{\text{WTA}}$ & $\mathcal{L}_{\text{SDF}}$ & $\mathcal{L}_{\text{thumb}}, \mathcal{L}_{\text{opp}}$ & $\mathcal{L}_{\text{closure}}$ & Naive Samp. & Rank. & TTP & $\text{SR}_{\text{scene}}$ & CFR \\
\midrule
\checkmark &            &            &            &            &            & \checkmark &            & 37.79 & 81.37 \\
           & \checkmark &            &            &            &            & \checkmark &            & 38.07 & 81.31 \\
           & \checkmark & \checkmark &            &            &            & \checkmark &            & 41.93 & 84.06 \\
           & \checkmark & \checkmark & \checkmark &            &            & \checkmark &            & 43.19 & 84.40 \\
           & \checkmark & \checkmark & \checkmark & \checkmark & \checkmark &            &            & 36.08 & 79.71  \\
           & \checkmark & \checkmark & \checkmark & \checkmark &            &            & \checkmark & 36.16 & 2.56 \\
           \rowcolor{gray!15} 
           & \checkmark & \checkmark & \checkmark & \checkmark &            & \checkmark &            & 44.41 & 83.66 \\

\bottomrule
\end{tabular}
}
\end{table}

\section{Conclusion}
\label{sec6}
In this paper, we presented OmniDex to tackle the data and algorithmic challenges of dexterous grasping in cluttered scenes. Our OmniDex benchmark leverages a scalable seed-and-filter strategy to deliver over 2.6 million scenes and 0.4 billion validated grasp ground truths, complete with rich multi-modal observations. Built upon this, our novel generation framework couples Soft Winner-Takes-All learning with human-inspired physical constraints to effectively capture grasp multimodality and bridge the crucial last-millimeter precision gap. Experimental results confirm that our OmniDex model achieves state-of-the-art performance on the proposed benchmark, significantly outperforming existing baselines on standard cluttered scenes.

\clearpage

{
    \small
    \bibliography{main}

@inproceedings{wang2023dexgraspnet,
  title={DexGraspNet: A Large-Scale Robotic Dexterous Grasp Dataset for General Objects Based on Simulation},
  author={Wang, Ruicheng and Zhang, Jialiang and Chen, Jiayi and Xu, Yinzhen and Li, Puhao and Liu, Tengyu and Wang, He},
  booktitle={IEEE Int. Conf. Robot. Autom.},
  pages={11359--11366},
  year={2023},
  organization={IEEE}
}

@inproceedings{wu2023omniobject3d,
  title={Omniobject3d: Large-vocabulary 3d object dataset for realistic perception, reconstruction and generation},
  author={Wu, Tong and Zhang, Jiarui and Fu, Xiao and Wang, Yuxin and Ren, Jiawei and Pan, Liang and Wu, Wayne and Yang, Lei and Wang, Jiaqi and Qian, Chen and others},
  booktitle={IEEE Conf. Comput. Vis. Pattern Recog.},
  pages={803--814},
  year={2023}
}

@article{chang2015shapenet,
  title={Shapenet: An information-rich 3d model repository},
  author={Chang, Angel X and Funkhouser, Thomas and Guibas, Leonidas and Hanrahan, Pat and Huang, Qixing and Li, Zimo and Savarese, Silvio and Savva, Manolis and Song, Shuran and Su, Hao and others},
  journal={arXiv preprint arXiv:1512.03012},
  year={2015}
}

@inproceedings{downs2022google,
  title={Google scanned objects: A high-quality dataset of 3d scanned household items},
  author={Downs, Laura and Francis, Anthony and Koenig, Nate and Kinman, Brandon and Hickman, Ryan and Reymann, Krista and McHugh, Thomas B and Vanhoucke, Vincent},
  booktitle={IEEE Int. Conf. Robot. Autom.},
  pages={2553--2560},
  year={2022},
}

@inproceedings{collins2022abo,
  title={Abo: Dataset and benchmarks for real-world 3d object understanding},
  author={Collins, Jasmine and Goel, Shubham and Deng, Kenan and Luthra, Achleshwar and Xu, Leon and Gundogdu, Erhan and Zhang, Xi and Vicente, Tomas F Yago and Dideriksen, Thomas and Arora, Himanshu and others},
  booktitle={IEEE Conf. Comput. Vis. Pattern Recog.},
  pages={21126--21136},
  year={2022}
}

@article{mittal2025isaac,
  title={Isaac lab: A gpu-accelerated simulation framework for multi-modal robot learning},
  author={Mittal, Mayank and Roth, Pascal and Tigue, James and Richard, Antoine and Zhang, Octi and Du, Peter and Serrano-Munoz, Antonio and Yao, Xinjie and Zurbr{\"u}gg, Ren{\'e} and Rudin, Nikita and others},
  journal={arXiv preprint arXiv:2511.04831},
  year={2025}
}

@inproceedings{zhang2024dexgraspnet,
  title={Dexgraspnet 2.0: Learning generative dexterous grasping in large-scale synthetic cluttered scenes},
  author={Zhang, Jialiang and Liu, Haoran and Li, Danshi and Yu, XinQiang and Geng, Haoran and Ding, Yufei and Chen, Jiayi and Wang, He},
  booktitle={Conf. Robot. Learn.},
  year={2024}
}

@inproceedings{chen2025clutterdexgrasp,
  title={ClutterDexGrasp: A Sim-to-Real System for General Dexterous Grasping in Cluttered Scenes},
  author={Chen, Zeyuan and Yan, Qiyang and Chen, Yuanpei and Wu, Tianhao and Zhang, Jiyao and Ding, Zihan and Li, Jinzhou and Yang, Yaodong and Dong, Hao},
  booktitle={Conf. Robot. Learn.},
  pages={885--905},
  year={2025},
  organization={PMLR}
}

@inproceedings{zhong2025dexgrasp,
  title={Dexgrasp anything: Towards universal robotic dexterous grasping with physics awareness},
  author={Zhong, Yiming and Jiang, Qi and Yu, Jingyi and Ma, Yuexin},
  booktitle={IEEE Conf. Comput. Vis. Pattern Recog.},
  pages={22584--22594},
  year={2025}
}

@inproceedings{wei2025afforddexgrasp,
  title={Afforddexgrasp: Open-set language-guided dexterous grasp with generalizable-instructive affordance},
  author={Wei, Yi-Lin and Lin, Mu and Lin, Yuhao and Jiang, Jian-Jian and Wu, Xiao-Ming and Zeng, Ling-An and Zheng, Wei-Shi},
  booktitle={Int. Conf. Comput. Vis.},
  pages={11818--11828},
  year={2025}
}

@inproceedings{yedex1b,
  title={Dex1B: Learning with 1B Demonstrations for Dexterous Manipulation},
  author={Ye, Jianglong and Wang, Keyi and Yuan, Chengjing and Yang, Ruihan and Li, Yiquan and Zhu, Jiyue and Qin, Yuzhe and Zou, Xueyan and Wang, Xiaolong},
  booktitle={RSS Workshop on Dexterous Manipulation},
  year={2025}
}

@inproceedings{he2025dexvlg,
  title={Dexvlg: Dexterous vision-language-grasp model at scale},
  author={He, Jiawei and Li, Danshi and Yu, Xinqiang and Qi, Zekun and Zhang, Wenyao and Chen, Jiayi and Zhang, Zhaoxiang and Zhang, Zhizheng and Yi, Li and Wang, He},
  booktitle={Int. Conf. Comput. Vis.},
  pages={14248--14258},
  year={2025}
}

@inproceedings{keselman2017intel,
  title={Intel realsense stereoscopic depth cameras},
  author={Keselman, Leonid and Iselin Woodfill, John and Grunnet-Jepsen, Anders and Bhowmik, Achintya},
  booktitle={IEEE Conf. Comput. Vis. Pattern Recog.},
  pages={1--10},
  year={2017}
}

@article{oquab2024dinov2,
  title={DINOv2: Learning Robust Visual Features without Supervision},
  author={Oquab, Maxime and Darcet, Timoth{\'e}e and Moutakanni, Th{\'e}o and Vo, Huy and Szafraniec, Marc and Khalidov, Vasil and Fernandez, Pierre and Haziza, Daniel and Massa, Francisco and El-Nouby, Alaaeldin and others},
  journal={Trans. Mach. Learn. Res.},
  year={2024}
}

@inproceedings{fang2020graspnet,
  title={Graspnet-1billion: A large-scale benchmark for general object grasping},
  author={Fang, Hao-Shu and Wang, Chenxi and Gou, Minghao and Lu, Cewu},
  booktitle={IEEE Conf. Comput. Vis. Pattern Recog.},
  pages={11444--11453},
  year={2020}
}

@article{fang2023anygrasp,
  title={Anygrasp: Robust and efficient grasp perception in spatial and temporal domains},
  author={Fang, Hao-Shu and Wang, Chenxi and Fang, Hongjie and Gou, Minghao and Liu, Jirong and Yan, Hengxu and Liu, Wenhai and Xie, Yichen and Lu, Cewu},
  journal={IEEE Trans. Robot.},
  volume={39},
  number={5},
  pages={3929--3945},
  year={2023},
}

@inproceedings{chendexonomy,
  title={Dexonomy: Synthesizing All Dexterous Grasp Types in a Grasp Taxonomy},
  author={Chen, Jiayi and Ke, Yubin and Peng, Lin and Wang, He},
  booktitle={RSS Workshop on Dexterous Manipulation},
  year={2025},
}

@inproceedings{liu2024realdex,
  title={RealDex: towards human-like grasping for robotic dexterous hand},
  author={Liu, Yumeng and Yang, Yaxun and Wang, Youzhuo and Wu, Xiaofei and Wang, Jiamin and Yao, Yichen and Schwertfeger, S{\"o}ren and Yang, Sibei and Wang, Wenping and Yu, Jingyi and others},
  booktitle={Proc. Int. Joint Conf. Artif. Intell.},
  pages={6859--6867},
  year={2024}
}

@article{duan2022survey,
  title={A survey of embodied ai: From simulators to research tasks},
  author={Duan, Jiafei and Yu, Samson and Tan, Hui Li and Zhu, Hongyuan and Tan, Cheston},
  journal={IEEE Trans. Emerg. Top. Comput. Intell.},
  volume={6},
  number={2},
  pages={230--244},
  year={2022},
}

@inproceedings{xu2023unidexgrasp,
  title={Unidexgrasp: Universal robotic dexterous grasping via learning diverse proposal generation and goal-conditioned policy},
  author={Xu, Yinzhen and Wan, Weikang and Zhang, Jialiang and Liu, Haoran and Shan, Zikang and Shen, Hao and Wang, Ruicheng and Geng, Haoran and Weng, Yijia and Chen, Jiayi and others},
  booktitle={IEEE Conf. Comput. Vis. Pattern Recog.},
  pages={4737--4746},
  year={2023}
}

@inproceedings{zhangcadgrasp,
  title={CADGrasp: Learning Contact and Collision Aware General Dexterous Grasping in Cluttered Scenes},
  author={Zhang, Jiyao and Ma, Zhiyuan and Wu, Tianhao and Chen, Zeyuan and Dong, Hao},
  booktitle={Adv. Neural Inf. Process. Syst.},
  year={2025}
}

@article{lundell2021ddgc,
  title={Ddgc: Generative deep dexterous grasping in clutter},
  author={Lundell, Jens and Verdoja, Francesco and Kyrki, Ville},
  journal={IEEE Robot. Autom. Lett.},
  volume={6},
  number={4},
  pages={6899--6906},
  year={2021},
}

@inproceedings{wang2025dexh2r,
  title={DexH2R: A benchmark for dynamic dexterous grasping in human-to-robot handover},
  author={Wang, Youzhuo and Ye, Jiayi and Xiao, Chuyang and Zhong, Yiming and Tao, Heng and Yu, Hang and Liu, Yumeng and Yu, Jingyi and Ma, Yuexin},
  booktitle={Int. Conf. Comput. Vis.},
  pages={12702--12712},
  year={2025}
}

@inproceedings{xudexumi,
  title={DexUMI: Using Human Hand as the Universal Manipulation Interface for Dexterous Manipulation},
  author={Xu, Mengda and Zhang, Han and Hou, Yifan and Xu, Zhenjia and Fan, Linxi and Veloso, Manuela and Song, Shuran},
  booktitle={Workshop on Human to Robot},
  year={2025}
}

@inproceedings{wan2023unidexgrasp++,
  title={Unidexgrasp++: Improving dexterous grasping policy learning via geometry-aware curriculum and iterative generalist-specialist learning},
  author={Wan, Weikang and Geng, Haoran and Liu, Yun and Shan, Zikang and Yang, Yaodong and Yi, Li and Wang, He},
  booktitle={Int. Conf. Comput. Vis.},
  pages={3891--3902},
  year={2023}
}

@inproceedings{hang2024dexfuncgrasp,
  title={Dexfuncgrasp: A robotic dexterous functional grasp dataset constructed from a cost-effective real-simulation annotation system},
  author={Hang, Jinglue and Lin, Xiangbo and Zhu, Tianqiang and Li, Xuanheng and Wu, Rina and Ma, Xiaohong and Sun, Yi},
  booktitle={AAAI},
  volume={38},
  number={9},
  pages={10306--10313},
  year={2024}
}

@inproceedings{qin2023dexpoint,
  title={Dexpoint: Generalizable point cloud reinforcement learning for sim-to-real dexterous manipulation},
  author={Qin, Yuzhe and Huang, Binghao and Yin, Zhao-Heng and Su, Hao and Wang, Xiaolong},
  booktitle={Conf. Robot. Learn.},
  pages={594--605},
  year={2023},
  organization={PMLR}
}

@inproceedings{chen2025bodex,
  title={Bodex: Scalable and efficient robotic dexterous grasp synthesis using bilevel optimization},
  author={Chen, Jiayi and Ke, Yubin and Wang, He},
  booktitle={IEEE Int. Conf. Robot. Autom.},
  pages={01--08},
  year={2025},
  organization={IEEE}
}

@article{loshchilov2017decoupled,
  title={Decoupled weight decay regularization},
  author={Loshchilov, Ilya and Hutter, Frank},
  journal={arXiv preprint arXiv:1711.05101},
  year={2017}
}

@misc{shadow_hand,
  author = {{Shadow Robot Company}},
  title = {Shadow Dexterous Hand},
  howpublished = {\url{https://www.shadowrobot.com/dexterous-hand/}},
  year = {2005}
}

@inproceedings{zhong2026dexgraspvla,
  title={Dexgraspvla: A vision-language-action framework towards general dexterous grasping},
  author={Zhong, Yifan and Huang, Xuchuan and Li, Ruochong and Zhang, Ceyao and Chen, Zhang and Guan, Tianrui and Zeng, Fanlian and Lui, Ka Nam and Ye, Yuyao and Liang, Yitao and others},
  booktitle={AAAI},
  volume={40},
  number={22},
  pages={18836--18844},
  year={2026}
}

@inproceedings{wei2024grasp,
  title={Grasp as you say: Language-guided dexterous grasp generation},
  author={Wei, Yi-Lin and Jiang, Jian-Jian and Xing, Chengyi and Tan, Xian-Tuo and Wu, Xiao-Ming and Li, Hao and Cutkosky, Mark and Zheng, Wei-Shi},
  journal={Adv. Neural Inf. Process. Syst.},
  volume={37},
  pages={46881--46907},
  year={2024}
}

@inproceedings{rupprecht2017learning,
  title={Learning in an uncertain world: Representing ambiguity through multiple hypotheses},
  author={Rupprecht, Christian and Laina, Iro and DiPietro, Robert and Baust, Maximilian and Tombari, Federico and Navab, Nassir and Hager, Gregory D},
  booktitle={Int. Conf. Comput. Vis.},
  pages={3591--3600},
  year={2017}
}

@misc{tianji_marvin_m6s,
  author = {{Tianji Intelligent Systems}},
  title = {Marvin M6S Long (CCS-920) Force-Controlled Humanoid Arm},
  note = {Product documentation},
  year = {2026}
}

@misc{brainco_revo2,
  author = {{BrainCo}},
  title = {Revo2 Bionic Dexterous Hand},
  note = {Product documentation},
  year = {2026}
}
}

\clearpage
\appendix

\section{Benchmark Generation Pipeline Details}
\label{AppA}
\setcounter{table}{0}
\renewcommand{\thetable}{A\arabic{table}}
\setcounter{figure}{0}
\renewcommand{\thefigure}{A\arabic{figure}}

\subsection{Cluttered Scene Geometry Database Generation}
\label{AppA.1}

\textbf{Supporting Base Asset Setting}. All supporting structures are canonicalized to a gravity-aligned orientation. We standardize their local coordinate frames by placing the origin at the intersection of the gravity axis and the ground plane, facilitating straightforward base anchoring within the simulation environment.

\textbf{Simulation Setup and Physical Properties}. We leverage Isaac Gym4 as our core physics engine to simulate clustered scene. To ensure stable and uniform dynamic behaviors across varied shapes, all object instances are assigned universally consistent physical properties: a density of 500 kg/m³, coupled with linear and angular damping factors of 1.0 and 0.3, respectively.

\textbf{Subset-Specific Initialization Strategies}. To systematically decouple the generalization evaluation, we employ tailored spatial initialization logic for each subset:

\textbullet $\mathcal{D}_{\text{std}}$ and $\mathcal{D}_{\text{mix}}$: We randomly sample 2 to 7 unique object instances. Objects are dropped from a height of 0.5m above the table and their dropping horizontal spawn coordinates are tightly bounded within the table's footprint.

\textbullet $\mathcal{D}_{\text{box}}$: 2 to 10 unique instances are initialized 0.4m above the top rim of the box. The horizontal dropping area is deliberately constricted to a 0.8 scale of the box's extent, naturally inducing dense and complex clustered stacking.

\textbullet $\mathcal{D}_{\text{grid}}$: We replicate 4 to 10 identical instances. To simulate structured arrays, objects are uniformly distributed within a 0.8 scaled tabletop area. Layouts are dynamically computed utilizing object bounding boxes, enforcing a strict minimum inter-instance spacing of 5cm.

\textbullet $\mathcal{D}_{\text{shelf}}$: Based on explicitly annotated layer heights, objects are uniformly arranged along each shelf layer using their bounding boxes, maintaining a minimum spacing of 4cm.

\textbf{Dynamic Resting Conditions}. The simulation evolves until the scene reaches a physically stable resting state. We define termination by two rigorous criteria: (1) kinematic energy dissipation, where the linear and angular velocities of all objects fall below 0.03 m/s and 0.10 rad/s, respectively; or (2) reaching a hard cutoff at 500 simulation steps to prevent infinite oscillation.

\textbf{Rigorous Geometric Validation}. To completely eliminate physically anomalous artifacts commonly found in automated generation pipelines, we implement a strict post-simulation geometric sanity check. A generated scene is systematically rejected unless it satisfies the following tolerance thresholds:

\textbullet Sinking \& Floating: The maximum allowed vertical deviation (embedding into the base or hovering above it) is strictly bounded to 3mm.

\textbullet Inter-object Penetration: The maximum permissible mutual penetration depth between any two clustered objects is capped at 2mm.

\textbf{Tolerance in Contact Filtering}. Our filtering is conservative but not perfectly zero-tolerance at the contact level. In practice, small numerical overlaps may arise from mesh decomposition, contact discretization, and simulator tolerances. We therefore allow a limited penetration margin during filtering and reject only clearly invalid grasp poses. In the current implementation, object--support penetration is tolerated up to 2cm before rejection, while shallower overlaps are retained to reduce false negatives caused by imperfect collision geometry.

\subsection{Cluttered Scene Visual Database Generation}
\label{AppA.2}

We design a multi-camera rig setup that mimics typical real-world robotic workspaces. We deploy five fixed Intel RealSense D455 sensors \cite{keselman2017intel}. To maximize the visual coverage of complex object interactions and mitigate severe occlusions, the camera poses are adaptively configured based on the subset:

\textbullet $\mathcal{D}_{\text{std}}$, $\mathcal{D}_{\text{mix}}$, $\mathcal{D}_{\text{grid}}$: For scenes without pronounced boundary occlusions, the primary ego-centric camera is positioned 0.6m above the global coordinate origin and retracted 0.3m outward from the table edge. Auxiliary multi-view cameras are symmetrically deployed 0.25m outward from the table corners. To guarantee optimal object framing, the optical axes of all cameras are explicitly oriented towards a focal point 0.15m below the spatial center.

\textbullet $\mathcal{D}_{\text{box}}$: The high physical boundaries of the container in the box subset introduce severe peripheral occlusions. To counteract this, we dynamically tighten the camera rig. The ego-centric camera and the auxiliary corner cameras are moved closer, fixed at a uniform 0.2m outward offset from the table edges and corners.

\textbullet $\mathcal{D}_{\text{shelf}}$: The multi-layered geometry of shelf imposes severe vertical visibility constraints, rendering standard top-down or steep perspectives entirely ineffective due to overhanging structures. We deploy a specialized 5-camera rig driven by a dynamic, layer-wise elevation-tracking strategy. The $z$-axis coordinates of all five cameras and their focal targets are aligned with the height of shelf, establishing a zero-pitch, horizon-aligned viewing frustum. The primary ego-centric camera is positioned squarely in front of the shelf, with its standoff distance adaptively scaled by a factor of 1.2 of the shelf's spatial diagonal. Simultaneously, the four auxiliary cameras are deployed radially outward from the top-surface corners, pushed along the diagonal axes by a scaling factor of 0.8.

\section{Benchmark Split and Density Details}
\label{AppB}
\setcounter{table}{0}
\renewcommand{\thetable}{B\arabic{table}}
\setcounter{figure}{0}
\renewcommand{\thefigure}{B\arabic{figure}}

To comprehensively assess grasping stability under varying degrees of physical entanglement, we deliberately modulate the cluster density, particularly for the $\mathcal{D}_{\text{std}}$ and $\mathcal{D}_{\text{mix}}$ subsets. This is explicitly achieved by constraining the object dropping area during simulation as a scalar ratio of the full tabletop extent. Specifically, we synthesize scenes across three distinct density tiers (ratios of 0.1, 0.2, and 0.3), with each density level contributing approximately 0.6M scenes to the overarching benchmark.

For the ablation mini-split, we uniformly sample $\sim$10\% of the training and testing scenes across all density tiers, yielding a diverse subset of roughly 90K scenes. This highly challenging spatial configuration allows us to rigorously benchmark the geometric reasoning capabilities of individual model components under dense clutter.

The asset pool is finite and fixed. To maximize the diversity of the training set, we intentionally assigned more challenging object assets to the seen objects split. Consequently, the remaining unseen Objects split is relatively simpler on average, and the two tracks are not difficulty-balanced.

\section{OmniDex Model Details}
\label{AppC}

\subsection{Single-View Condition Embedding}
\label{AppC.1}

For the ego-centric single-view setting, we extract patch-level semantic features from the RGB image using a pre-trained DINOv2 \cite{oquab2024dinov2} backbone, yielding ${{{\bf{F}}_{\rm{img}}} \in {{ \mathbb{R} }^{N \times d}}}$, where ${N}$ is the number of patches and ${d}$ is the embedding dimension. To encode scene geometry, we back-project the depth map into a 3D coordinate map ${{\bf{X}} \in {{ \mathbb{R} }^{H \times W \times 3}}}$:
\begin{equation}
\label{eq1}
{\bf{X}}\left( {u,v} \right) = {\bf{D}}\left( {u,v} \right){{\bf{K}}^{ - 1}}{\left[ {u,v,1} \right]^T},
\end{equation}
where ${\left( {u,v} \right)}$ denotes the 2D pixel coordinate. We then use a lightweight adapter to map the XYZ representation into the DINOv2 input space, producing geometric features ${{{\bf{F}}_{\rm{depth}}} \in {{ \mathbb{R} }^{N \times d}}}$.

To isolate the target object from clutter, the instruction mask is encoded into mask features ${{{\bf{F}}_{\rm{mask}}} \in {{ \mathbb{R} }^{N \times d}}}$. We concatenate semantic, geometric, and mask features along the channel dimension and fuse them into a scene-aware representation ${{{\bf{F}}_{\rm{scene}}} \in {{ \mathbb{R} }^{N \times d}}}$.

To preserve a consistent global reference for grasp generation, we additionally encode the camera parameters and table height. Specifically, we compute the projection matrix ${{\bf{P}} = {\bf{KT}}}$ and encode ${{\bf{P}}}$ together with ${{\bf{H}}}$ via MLPs to obtain a parameter token ${{{\bf{f}}_{{\rm{para}}}} \in {{ \mathbb{R} }^{1 \times d}}}$. Appending ${{{\bf{f}}_{{\rm{para}}}}}$ to ${{{\bf{F}}_{{\rm{scene}}}}}$ yields the final condition embedding ${{\bf{c}} \in {{ \mathbb{R} }^{\left( {N + 1} \right) \times d}}}$.

\subsection{Multi-View Input Fusion}
\label{AppC.2}

To effectively process extreme spatial occlusions, we introduce a hybrid and physics-aware encoder for multi-view observation. Instead of relying on a monolithic black-box transformer to indiscriminately mix multi-view input, our architecture deliberately decouples the fusion paradigm into two distinct pathways: a data-driven semantic early RGB fusion, a rigorous geometry-aware depth reprojection, and camera relationship tokens.

\textbf{Data-Driven Early RGB Fusion}. Traditional multi-view encoders often suffer from computational overhead when processing independent visual tokens across views. To ensure efficiency without losing critical textural semantics, we concatenate the multi-view RGB, depth, and instance masks, feeding them into a lightweight convolutional adapter. This yields an inter-view alpha map via a softmax operation across the view dimension $V$. The RGB inputs are then aggregated using this learned alpha map, effectively allowing the network to dynamically stitch unoccluded visual patches from auxiliary cameras into a cohesive representation.

\textbf{Rigorous Geometry-Aware Depth Reprojection}. Unlike textural features, depth represents strict spatial geometry that cannot be naively interpolated. To capture the 3D manifold of the cluttered scene, we explicitly route the multi-view depth maps to the primary anchor camera using strict camera parameters. To maintain computational tractability, this point-cloud inversion is performed utilizing a hardware-accelerated Z-buffer scattering mechanism. The physically fused depth map is subsequently converted into a normalized XYZ spatial map. 

\textbf{Camera Relationship Tokens}. To supplement the dense visual features, our encoder explicitly injects global contextual priors. The camera poses are tokenized and processed through a View Relation Transformer to capture the spatial topology of the multi-camera rig. These auxiliary tokens are concatenated with the visual features to form the final comprehensive observation sequence for the downstream generative policy.

\subsection{Physically-Constrained Loss}
\label{AppC.3}
We first back-project the depth map and filter it utilizing the instruction mask to extract a partial 3D point cloud of the targeted object, denoted as ${{{\cal P}_{{\rm{obj}}}}}$. Simultaneously, we obtain the predicted grasp pose ${\widehat {\bf{g}}}$ from the denoised output and feed it into the differentiable dexterous hand simulator to compute the predefined contact points on the hand surface \cite{wang2023dexgraspnet}, denoted as ${{{\cal P}_{{\rm{con}}}}}$. For any point ${p \in {{\cal P}_{{\rm{con}}}}}$, its minimal distance to the object point cloud is defined as $d(p) = \min_{q \in \mathcal{P}_{\text{obj}}} \|p - q\|_2$.

Effective grasping strictly requires a stable kinematic clamping force. Morphologically, the thumb plays a dominant role in opposing the other fingers. We explicitly enforce this by partitioning the contact points into a thumb set ${{{\cal P}_{{\rm{thumb}}}}}$ and an opposing fingers set ${{{\cal P}_{{\rm{opp}}}}}$. Specifically, we apply a thumb contact loss as:

\begin{equation}
\label{eq5}
\mathcal{L}_{\text{thumb}} = \max \left( \frac{1}{|\mathcal{P}_{\text{thumb}}|} \sum_{p \in \mathcal{P}_{\text{thumb}}} d(p) - \epsilon_{\text{contact}}, 0 \right)
\end{equation}

\noindent where $\epsilon_{\text{contact}}$ is a contact tolerance margin.

For the opposing fingers, forming a stable force closure often only requires a partial subset to make contact. Therefore, we apply an identical loss formulation for $\mathcal{L}_{\text{opp}}$, but compute the average distance over exclusively the $K$ closest points within $\mathcal{P}_{\text{opp}}$. 

Beyond simple proximity, a stable grasp necessitates a valid force closure, requiring the fingers to approach the object from opposing directions. To explicitly encourage this antipodal grasping pattern, we introduce a force closure loss as:

\begin{equation}
\label{eq6}
{{{\cal L}_{{\rm{closure}}}} = \max \left( {\cos \left( {{v_{{\rm{thumb}}}},{v_{{\rm{opp}}}}} \right) + \epsilon_{\text{closure}},0} \right)}
\end{equation}

where ${{v_{{\rm{thumb}}}},{v_{{\rm{opp}}}}}$ are approach vectors from finger to object center, $\cos(\cdot, \cdot)$ denotes the cosine similarity, and $\epsilon_{\text{closure}}$ is a tolerance margin.

Finally, to prevent physically impossible interpenetrations between the hand and the target object, we incorporate a Signed Distance Field (SDF) penetration loss ${{{\cal L}_{{\rm{SDF}}}}}$. Let $\Phi_{\text{obj}}(\cdot)$ denote the SDF of the object, where negative values indicate spatial locations strictly inside the object's geometry. We penalize any hand mesh vertices ${{{\cal P}_{{\rm{hand}}}}}$ that penetrate the object as follows:

\begin{equation}
\label{eq7}
\mathcal{L}_{\text{SDF}} = \frac{1}{|\mathcal{P}_{\text{hand}}|} \sum_{p \in \mathcal{P}_{\text{hand}}} \max(-\Phi_{\text{obj}}(p), 0)
\end{equation}

\noindent where $\mathcal{P}_{\text{hand}}$ represents the set of all vertices on the predicted hand mesh. This term effectively repels the hand fingers and palm from intersecting with the object's interior volume.

\section{Supplementary Implementation Details}
\label{AppD}
\setcounter{table}{0}
\renewcommand{\thetable}{C\arabic{table}}
\setcounter{figure}{0}
\renewcommand{\thefigure}{C\arabic{figure}}

All experiments are conducted on NVIDIA A800 GPUs. The generation of the cluster scene geometry database and the cluster scene visual database consumed about 5K and 17K GPU hours, respectively. The AdamW optimizer \cite{loshchilov2017decoupled} is used with an initial learning rate of ${{e^{ - 4}}}$ and a weight decay of ${{e^{ - 4}}}$, All training durations are set to 25 epochs. For both the STA loss optimization during training and the physics-driven ranking strategy during inference, we uniformly set the number of parallel grasp hypotheses to $M = 64$. We define loss weights $\lambda_{\text{WTA}} = 1$ and $\lambda_{\text{phys}} = 100$, respectively. Within the physical constraint formulation, the relative weighting among the sub-components is specifically calibrated as $\mathcal{L}_{\text{thumb}} : \mathcal{L}_{\text{opp}} : \mathcal{L}_{\text{SDF}} : \mathcal{L}_{\text{closure}} = 5 : 1 : 0.2 : 2$. Notably, we assign a substantially higher penalty weight to the thumb, explicitly reflecting its indispensable kinematic role in establishing robust force closure. Imposing strict physical penalties on an uninitialized network often leads to severe learning instability. To mitigate this, we employ a delayed linear warm-up scheduling for the physical constraints. Specifically, the physical loss term is entirely deactivated ($\lambda_{\text{phys}} = 0$) during the initial 10 epochs, allowing the generative network to first learn a stable grasp manifold. Subsequently, $\lambda_{\text{phys}}$ is linearly ramped up to its maximum target value of 100 between epochs 10 and 20, ensuring a smooth transition towards physically compliant grasp synthesis.

Our benchmark is constructed from third-party assets obtained from their official releases, including OmniObject3D \cite{wu2023omniobject3d}, ShapeNetCore v2 \cite{chang2015shapenet}, Google Scanned Objects \cite{downs2022google}, ABO \cite{collins2022abo}, and Isaac Lab \cite{mittal2025isaac}. We credit these original resources in the paper and use them in accordance with their respective licenses and terms of use. In particular, OmniObject3D and Google Scanned Objects are released under CC BY 4.0, ABO is released under CC BY 4.0, ShapeNetCore v2 is used under the ShapeNet Terms of Use for non-commercial research, and Isaac Lab is distributed under the BSD-3-Clause license.

\section{Supplementary Ablation Study}
\label{AppE}
\setcounter{table}{0}
\renewcommand{\thetable}{D\arabic{table}}
\setcounter{figure}{0}
\renewcommand{\thefigure}{D\arabic{figure}}

\subsection{Inference Ablation}
\label{AppE.1}

To evaluate the efficiency-performance trade-off during inference, we ablate $M$, the number of parallel generated grasp hypotheses, and comparison with TTP. For a rigorously fair comparison, the TTP baseline is configured to optimize the exact same physical loss objectives utilized in the training phase, executing for 200 steps per grasp. Quantitative results are detailed in Table \ref{tableB}.

\begin{table}[htbp]
\centering
\caption{Effect of parallel hypothesis generation vs. Test-Time Optimization.}
\resizebox{\linewidth}{!}{
\label{tableB}
\begin{tabular}{lcccccccc}
\toprule
\multirow{2}{*}{Metric} & \multicolumn{7}{c}{Parallel Hypotheses ($M$)} & \multirow{2}{*}{TTP} \\
\cmidrule(lr){2-8}
& 1 & 2 & 4 & 8 & 16 & 32 & 64 \\
\midrule
$\text{SR}_{\text{scene}}$ & 36.08&	38.02&	41.06&	42.47&	43.41&	44.37&	44.41&	36.16\\
CFR & 79.71	&80.81	&81.58&	82.26&	82.62&	83.09&	83.66&	2.56 \\
Inference Time (ms) & 219.14	&233.35&	266.39&	325.20&	543.93&	948.33&	1598.41&	6044.60  \\
\bottomrule
\end{tabular}
}
\end{table}

As observed, scaling the hypothesis pool $M$ consistently improves grasping success rates with only a negligible sacrifice in inference latency, validating the high efficiency of our parallel generation paradigm. In stark contrast, TTP incurs prohibitive computational overhead and suffers from severe optimization instability. We empirically find that the TTP trajectory frequently diverges from physically viable grasp manifolds, ultimately degrading both execution speed and overall performance.

\subsection{Failure Analysis}
\label{AppE.2}

We analyze OmniDex failures on the $\mathcal{D}_{\text{std}}$ mini-split. As summarized in Table~\ref{tableC}, among grasps failing $\text{SR}_{\text{scene}}$, 4.29\% are caused by insufficient contact and 95.71\% by stability failure after contact is established. Among grasps failing CFR, 61.26\% involve supporting-base collisions and 38.74\% involve non-target-object collisions. These results indicate that post-contact grasp stability and collision avoidance in clutter are the dominant remaining challenges.

\begin{table}[htbp]
\centering
\caption{Failure breakdown on the $\mathcal{D}_{\text{std}}$ mini-split. }
\label{tableC}
\small
\begin{tabular}{lclc}
\toprule
\multicolumn{2}{c}{$\text{SR}_{\text{scene}}$ failures} &
\multicolumn{2}{c}{CFR failures} \\
\cmidrule(lr){1-2} \cmidrule(lr){3-4}
Failure type & Rate & Failure type & Rate \\
\midrule
Insufficient contact & 4.29\% &
Supporting-base collision & 61.26\% \\
Stability failure & 95.71\% &
Non-target-object collision & 38.74\% \\
\bottomrule
\end{tabular}
\end{table}

\subsection{IK Feasibility Audit}
\label{AppE.3}

To assess the kinematic feasibility of successful predicted grasps, we evaluate the $\mathcal{D}_{\text{std}}$ mini-split using a Franka arm placed on the ego-view side of the table. We consider five representative arm-base configurations, where the rear offset denotes the horizontal distance from the table edge and the height offset denotes the vertical displacement above the tabletop. For each grasp passing $\text{SR}_{\text{scene}}$, we transform the end-effector pose into the arm-base frame and solve IK subject to the Franka joint limits. IK feasibility ranges from 83.75\% to 87.98\% across the five configurations, indicating that most generated grasps are kinematically reachable by the robot arm. The detailed results are shown in Table~\ref{tableD}.

\begin{table}[htbp]
\centering
\caption{IK feasibility under five representative Franka arm placements on the $\mathcal{D}_{\text{std}}$ mini-split.}
\label{tableD}
\small
\begin{tabular}{ccc}
\toprule
Rear offset (cm) & Height offset (cm) & IK feasibility (\%) \\
\midrule
0.0 & +2.5 & 87.98 \\
0.0 & +5.0 & 87.94 \\
2.5 & +5.0 & 86.68 \\
5.0 & +2.5 & 85.63 \\
7.5 & +5.0 & 83.75 \\
\bottomrule
\end{tabular}
\end{table}

\section{Supplementary Visual Results}
\label{AppF}
\setcounter{table}{0}
\renewcommand{\thetable}{E\arabic{table}}
\setcounter{figure}{0}
\renewcommand{\thefigure}{E\arabic{figure}}

\subsection{OmniDex Benchmark Visual Results}
\label{AppF.1}

\begin{figure*}[!htbp] 
\begin{center}
\includegraphics[width=\linewidth]{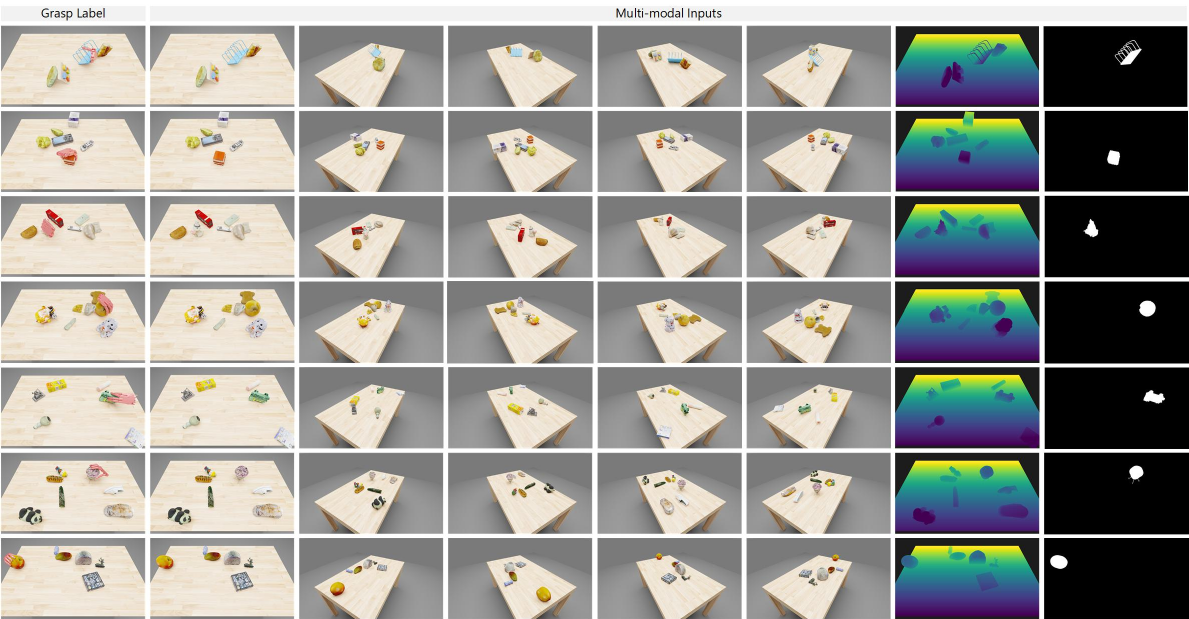}  
\end{center}
\caption{Qualitative visualization of the $\mathcal{D}_{\text{std}}$ subset.}
\label{fige1}
\end{figure*}

\begin{figure*}[!htbp] 
\begin{center}
\includegraphics[width=\linewidth]{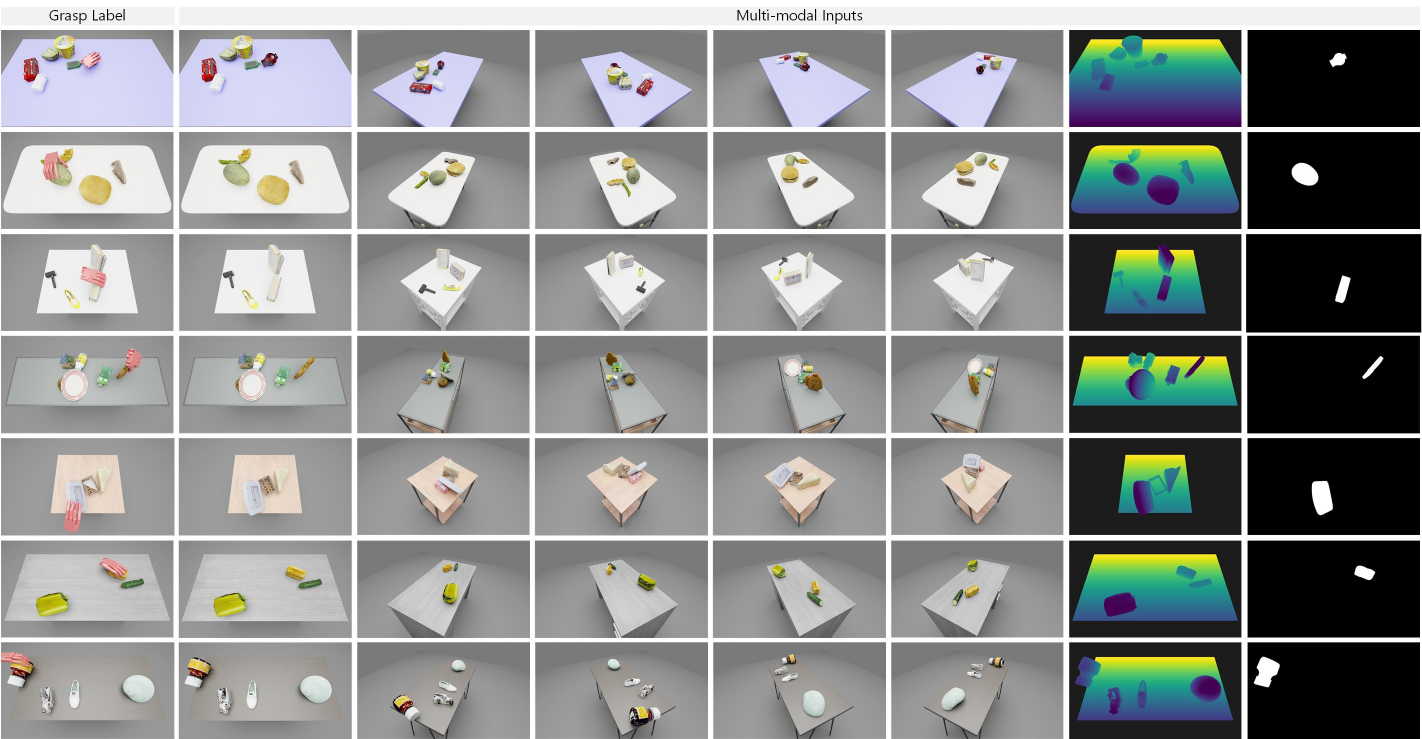}  
\end{center}
\caption{Qualitative visualization of the $\mathcal{D}_{\text{mix}}$ subset.}
\label{fige2}
\end{figure*}

\begin{figure*}[!htbp] 
\begin{center}
\includegraphics[width=\linewidth]{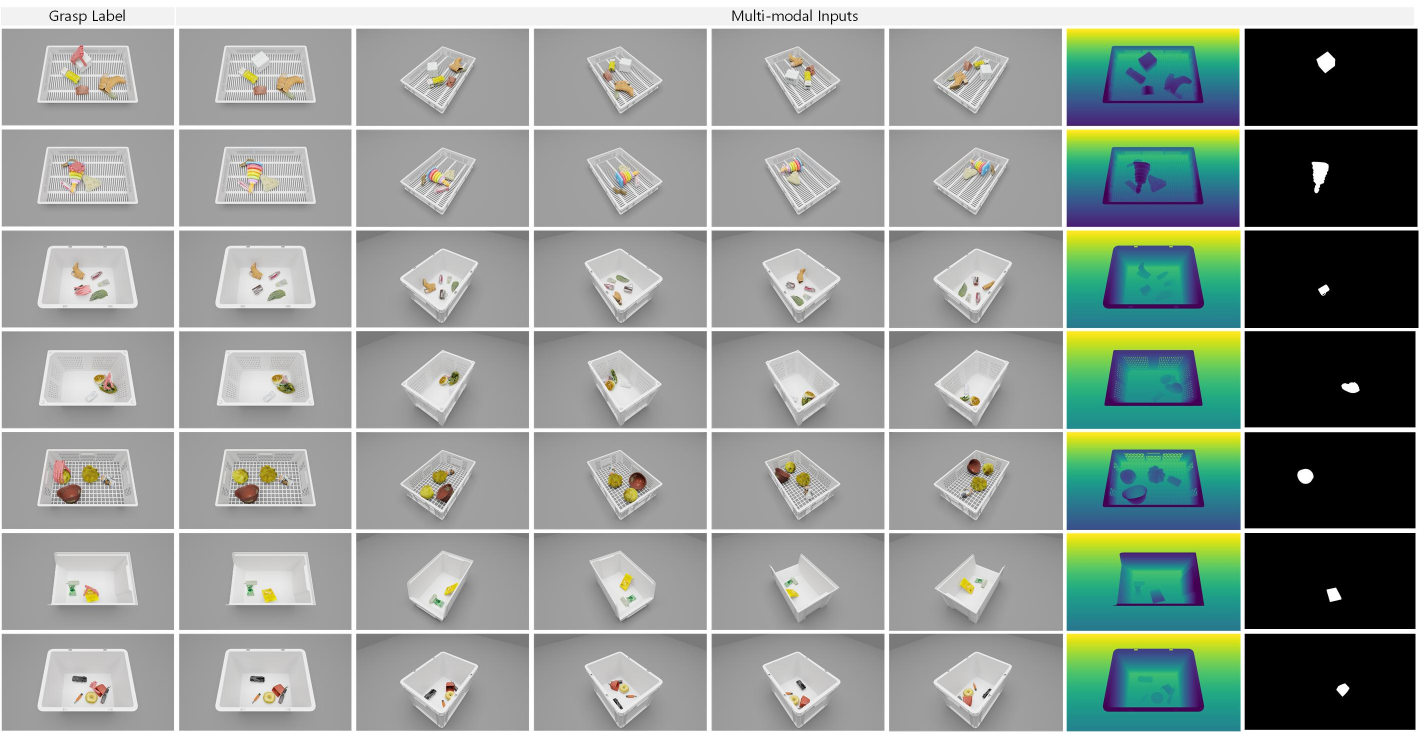}  
\end{center}
\caption{Qualitative visualization of the $\mathcal{D}_{\text{box}}$ subset.}
\label{fige3}
\end{figure*}

\begin{figure*}[!htbp] 
\begin{center}
\includegraphics[width=\linewidth]{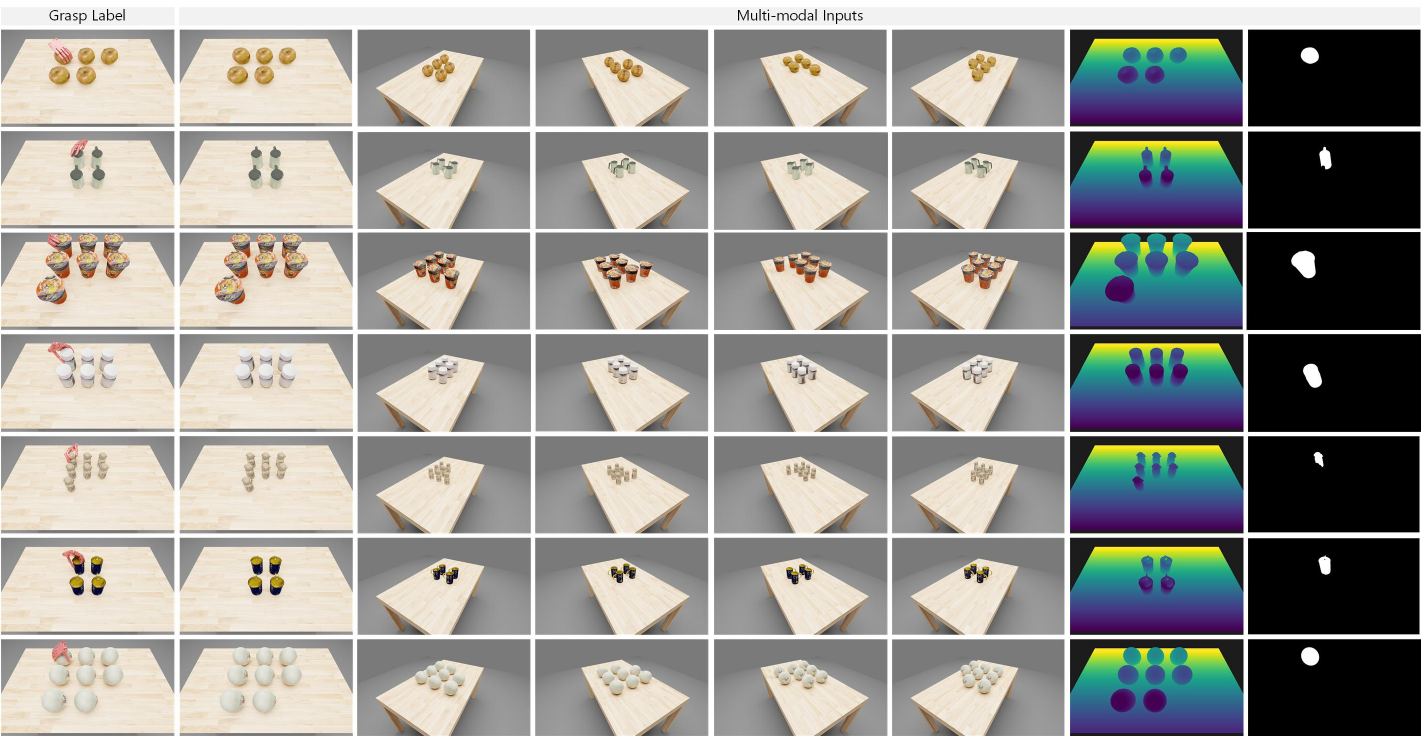}  
\end{center}
\caption{Qualitative visualization of the $\mathcal{D}_{\text{grid}}$ subset.}
\label{fige4}
\end{figure*}

\begin{figure*}[!htbp] 
\begin{center}
\includegraphics[width=\linewidth]{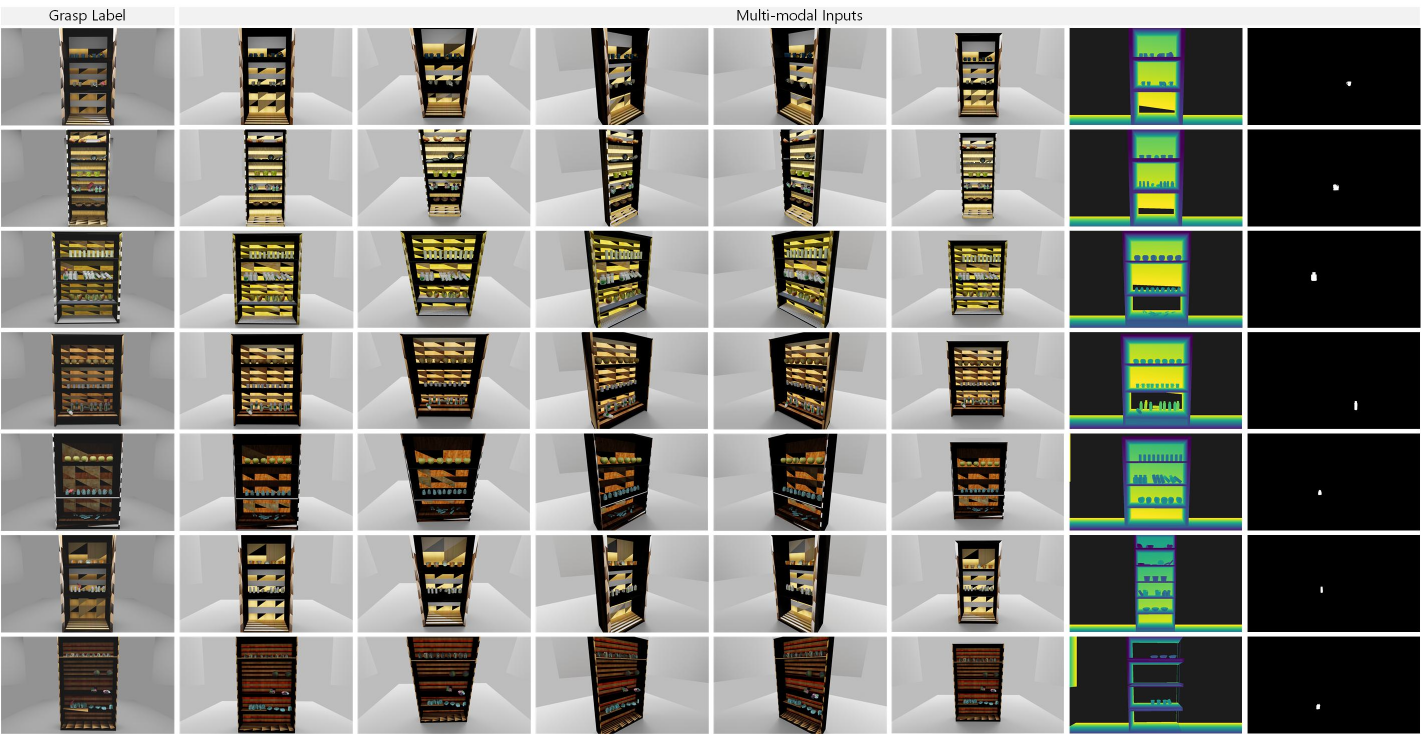}  
\end{center}
\caption{Qualitative visualization of the $\mathcal{D}_{\text{shelf}}$ subset.}
\label{fige5}
\end{figure*}

To comprehensively demonstrate the high fidelity and multi-modal richness of our proposed OmniDex benchmark, we present extensive visual galleries for each subset in Figures \ref{fige1} through \ref{fige5}. For every representative scene, we visualize the complete suite of aligned multi-modal sensory inputs: multi-view camera images, depth maps, instruction masks, and the physically-validated dexterous grasp ground truths. These visualizations collectively underscore the extreme spatial complexity of the cluttered layouts and the rigorous alignment of our grasping label across diverse visual modalities.

\subsection{OmniDex Model Visual Results}
\label{AppF.2}

\begin{figure*}[!htbp] 
\begin{center}
\includegraphics[width=\linewidth]{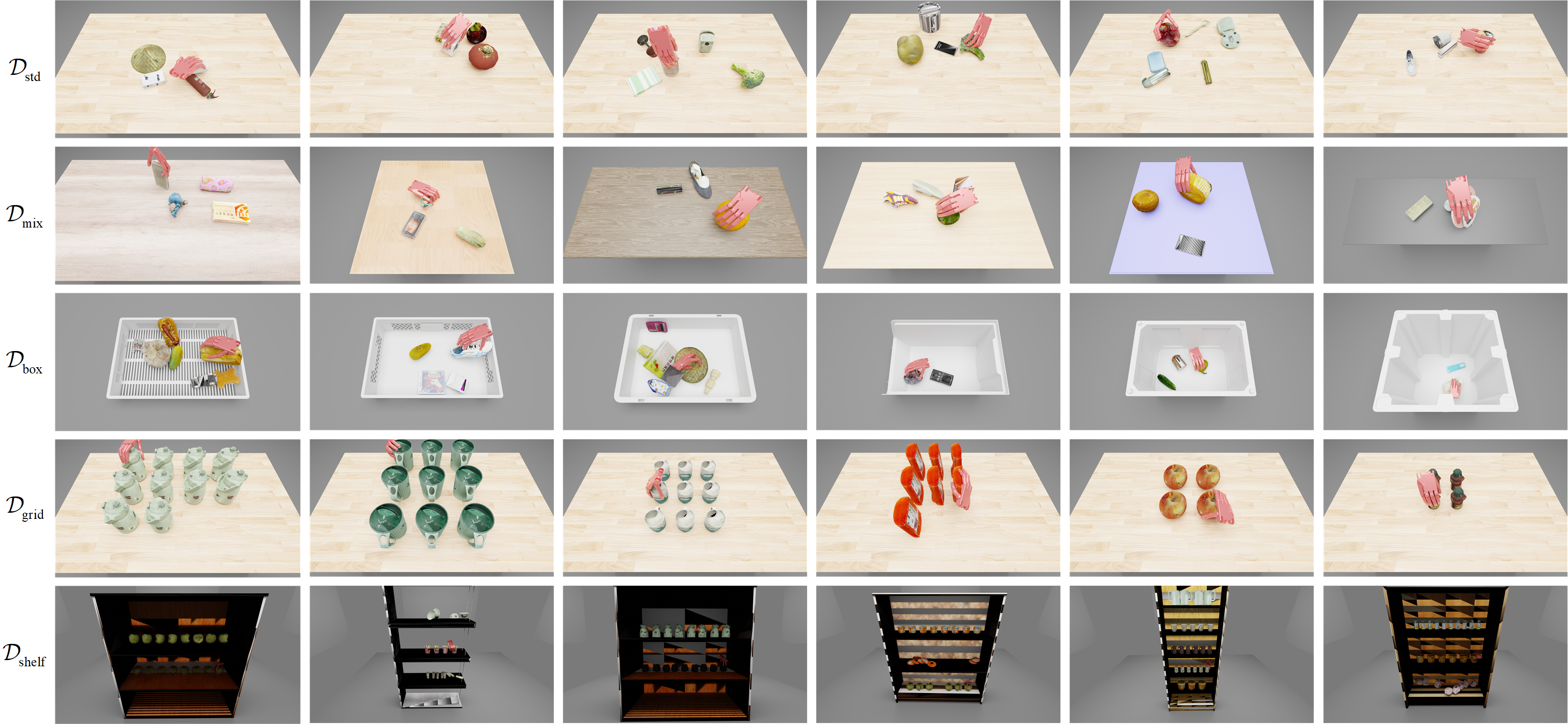}  
\end{center}
\caption{Qualitative results of OmniDex model across diverse benchmark subsets.}

\label{fig6}
\end{figure*}

We present qualitative results of OmniDex model across diverse benchmark subsets in Fig.~\ref{fig6}. The predicted grasps remain stable and semantically consistent under substantially different scene layouts. Notably, OmniDex is able to localize the target object and generate plausible grasping poses even in the presence of tight inter-object spacing and strong geometric ambiguity among similar instances. These visualizations further support our quantitative findings, showing that the proposed multimodal conditioning and physics-aware generation scheme generalize well across varying scene structures and grasping difficulties.

\subsection{Real-Robot Demonstration}
\label{AppF.3}

\begin{figure*}[!htbp]
\centering
\includegraphics[width=0.8\linewidth]{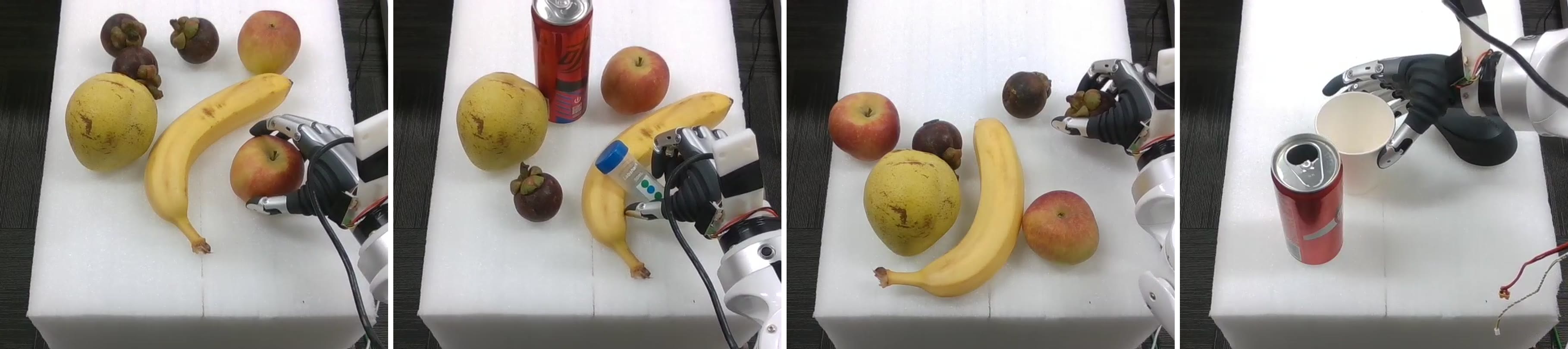}
\caption{Representative real-robot grasping trials in cluttered scenes.}
\label{fig:real_robot}
\end{figure*}

We conducted a small qualitative real-robot study using a Tianji Marvin M6S arm~\cite{tianji_marvin_m6s} and a BrainCo Revo2 hand~\cite{brainco_revo2}, with the camera placement approximately matching the simulated setup. OmniDex was retrained on a hardware-matched mini-split, and SAM generated the target masks. As shown in Fig.~\ref{fig:real_robot}, OmniDex conducted representative grasping trials on an apple, a mangosteen, and a cup in cluttered scenes.

\section{Limitation}
\label{AppG}
Despite the scale and diversity of the proposed benchmark, both the benchmark construction and the main evaluation are conducted in simulation. Although we incorporate diverse supporting bases, multi-view RGB-D observations, and physics-based validation to better approximate realistic cluttered grasping, a non-negligible sim-to-real gap may still remain due to unmodeled factors such as sensor noise, calibration error, contact parameter mismatch, and actuation uncertainty. While our seed-and-filter pipeline is substantially more scalable than scene-level optimization and our inference is more efficient than test-time optimization, the overall system still requires considerable computational resources for large-scale data generation and model training, constructing the full benchmark and training the model involve substantial GPU cost. Addressing real-world transfer and further reducing the resource requirement are important directions for future work.


\end{document}